\documentclass{article}

\usepackage{arxiv}

\usepackage[utf8]{inputenc} 
\usepackage[T1]{fontenc}    
\usepackage{hyperref}       
\usepackage{url}            
\usepackage{booktabs}       
\usepackage{amsfonts}       
\usepackage{nicefrac}       
\usepackage{microtype}      
\usepackage{lipsum}		
\usepackage{graphicx}
\usepackage{natbib}
\usepackage{doi}
\usepackage{xcolor}
\usepackage{amsmath}
\usepackage{physics}
\usepackage{bm}
\usepackage{algorithm}
\usepackage{algpseudocode}
\floatname{algorithm}{Pseudocode}
\usepackage{arydshln}
\usepackage{caption}
\usepackage{subcaption}

\title{Using dynamic loss weighting to boost improvements \\ in forecast stability}

\date{January 20, 2025}	

\author{Daan Caljon\\
	KU Leuven\\
	\And
	Jeff Vercauteren\\
	KU Leuven\\
	\And
        Simon De Vos\\
        KU Leuven\\
	\And
        Wouter Verbeke\\
        KU Leuven\\
	\And
	Jente Van Belle\thanks{Corresponding author: \texttt{jente.vanbelle@kuleuven.be}.}\\
        KU Leuven\\
}

\renewcommand{\shorttitle}{Using dynamic loss weighting to boost improvements in forecast stability}

\hypersetup{
pdftitle={Using dynamic loss weighting to boost improvements in forecast stability},
pdfsubject={cs.LG, stat.ML, stat.ME},
pdfauthor={Daan Caljon, Jeff Vercauteren, Simon De Vos, Wouter Verbeke, Jente Van Belle},
pdfkeywords={Deep learning, Dynamic hyperparameter tuning, Rolling origin forecast instability, Global models, N-BEATS},
}

\begin{document}
\maketitle

\begin{abstract} 
Rolling origin forecast instability refers to variability in forecasts for a specific period induced by updating the forecast when new data points become available. 
Recently, an extension to the N-BEATS model for univariate time series point forecasting was proposed to include forecast stability as an additional optimization objective, next to accuracy.
It was shown that more stable forecasts can be obtained without harming accuracy by minimizing a composite loss function that contains both a forecast error and a forecast instability component, with a static hyperparameter to control the impact of stability.
In this paper, we empirically investigate whether further improvements in stability can be obtained without compromising accuracy by applying dynamic loss weighting algorithms, which change the loss weights during training.
\textcolor{black}{We show that 
existing dynamic loss weighting methods
can
achieve this objective and provide insights into why this might be the case.
Additionally, we propose an extension to the Random Weighting approach---Task-Aware Random Weighting---which also achieves this objective.}
\vspace{0.5cm}
\end{abstract}


\setcounter{footnote}{0} 



\section{Introduction} \label{SEC:INTRODUCTION}

In practice, multi-step-ahead forecasts are often updated when new observations become available (i.e., when time passes).
The underlying idea is that forecasts
typically
improve in accuracy
as the target time period approaches.
However, at the same time, updating forecasts can 
lead to substantial adjustments to earlier predictions for those same periods.
\citet{vanbelle2023} refer to these changes as \textit{rolling origin forecast instability} and define it as ``the variability in forecasts for a specific time period caused by updating the forecast for this time period each time a new observation becomes available, or in other words, from using subsequent forecasting origins (i.e., the time period from which the forecast is generated)" (p.\ 1334).
Depending on the forecasting method used, forecast instability can stem from the impact of (a) newly available observation(s) on either parameter estimation alone or on both model selection and parameter estimation.
Hereafter, with the term forecast (in)stability, we specifically refer to rolling origin forecast (in)stability.

If forecasts are used as a means to an end in that they are used as inputs to draw up plans, forecast updates give rise to both benefits and costs. 
Using 
more accurate
updated forecasts as inputs to plans, after all, makes sure that the plan is more closely aligned with what eventually will happen; however, the induced forecast instabilities may 
also 
lead to 
costly 
adjustments
to these plans.
For example, in a supply chain planning context,
forecast instabilities can result in costs
due to necessary revisions of initial supply plans,
which may cause excess inventory build-up or require expedited production and/or delivery
\citep{tunc2013simple,li2017}.

If forecasts are not updated, we avoid the additional costs from induced forecast instability; 
however, we also miss out on the potential benefits resulting from improvements in forecast accuracy.
Ideally, if we could fully quantify these associated costs and benefits, we could optimally trade off accuracy and stability. However, this quantification is often difficult in practice \citep{tunc2013simple}. An alternative approach is to focus on improving stability without sacrificing accuracy \citep{vanbelle2023}.
This approach is reasonable because, when forecasts are updated in practice, it is implicitly assumed that the benefits of improved accuracy outweigh the costs of induced instability.
Nevertheless, (slightly) less accurate but more stable algorithmic forecasts might be justified, as 
instabilities can cause (non-technical) users to lose trust in the forecasting system, possibly resulting in unwarranted judgmental adjustments that might reduce forecast accuracy \citep{petropoulos2022forecasting}.

\citet{vanbelle2023} propose a methodology for optimizing global neural point forecasting models for both forecast accuracy and stability
that has been empirically shown to improve forecast stability 
without leading to considerable losses in accuracy.
\textcolor{black}{The key element of their proposed solution is to cast the problem as a bi-objective optimization task by using a composite loss function,}
which can be conceptually formulated as follows:
\vspace{0.15cm}
\begin{equation} \label{EQ:L_COMBINED}
\mathcal{L}_\text{combined} = (1-\lambda) \cdot \mathcal{L}_\text{error}(\hat{y}, y) + \lambda \cdot \mathcal{L}_\text{instability}(\hat{y}, \hat{y}_\text{old}),
\end{equation}
where $\mathcal{L}_\text{error}$ is a loss function that optimizes forecast accuracy 
by quantifying the difference between the forecast $\hat{y}$ and the observed value $y$,
$\mathcal{L}_\text{instability}$ is a loss function for forecast stability
that measures the difference between the forecast $\hat{y}$ and an older forecast $\hat{y}_\text{old}$ for the same time period, and $\lambda \in [0,1]$ is a hyperparameter that controls the weight assigned to forecast instability during training.
They apply this methodology to extend the 
N-BEATS \citep{oreshkin2020nbeats} deep learning method for univariate point forecasting, resulting in a new method named N-BEATS-S.
The authors empirically demonstrate that there are $\lambda$ values 
for which N-BEATS-S produces 
\textcolor{black}{forecasts that are as accurate but more stable than those of N-BEATS.}
Moreover, in some cases, N-BEATS-S even outperforms N-BEATS in terms of forecast accuracy, suggesting that 
the forecast instability loss term
may also serve as a regularization mechanism,
potentially improving generalization performance.

\textcolor{black}{From a machine learning perspective,
if optimizing for forecast accuracy and stability is framed as two distinct tasks,
the methodology described above can be viewed as a multi-task learning (MTL) problem.}
In MTL, the goal is to improve generalization performance by learning multiple tasks in parallel while using a shared representation \citep{caruana1997multitask}. 
As stated by \citet{caruana1997multitask}, ``when alternate metrics [...] capture different, but useful, aspects of a problem, MTL can be used to benefit from them" (p.\ 56).
While the MTL literature is not limited to work on neural networks only \citep{wei2022mtl}, the majority of the work focuses on MTL with deep learning.
Traditionally, MTL networks are trained using a composite loss function that combines the losses of separate tasks with static coefficients.
In \citet{vanbelle2023}, a static value for $\lambda$ in Equation~\eqref{EQ:L_COMBINED} is determined by using grid search.
This approach thus coincides with the traditional method for training MTL networks.
More recently, however, much of the work in the deep MTL field has focused on leveraging the iterative nature of optimizing neural networks
\textcolor{black}{to address the challenge of optimizing multiple objectives simultaneously}
by dynamically changing the loss weights during training (i.e., different combination weights are used in different learning iterations).
We \textcolor{black}{conjecture} that 
\textcolor{black}{applying}
dynamic loss weighting (DLW) algorithms could 
improve the performance of N-BEATS-S for two reasons.
First, DLW methods have been demonstrated to enhance performance on other MTL problems \textcolor{black}{\citep[see, e.g.,][]{chen2018gradnorm,kendall2018multi,yu2020gradient,lin2022reasonable,navon2022multi,verboven2022hydalearn}}.
Second, perfect forecast stability is straightforward to achieve because a model may simply learn to predict the same constant value (e.g., zero) for all inputs, resulting in perfectly stable but most likely very inaccurate forecasts. If the model is trained with a fixed and relatively high $\lambda$, it might get stuck in a local optimum, achieving poor accuracy but near-perfect stability. When using grid search to select a static value for $\lambda$, the procedure might favor lower $\lambda$ values to avoid these local optima, introducing a bias toward selecting a relatively low $\lambda$. We hypothesize that dynamically tuning $\lambda$ to prioritize forecast accuracy during the early stages of training, and only increasing $\lambda$ after this initial phase---keeping in mind the goal of improving forecast stability without sacrificing accuracy---can lead to improved performance compared to using a tuned static value for $\lambda$.

In this paper, we explore the potential of dynamically weighting the two components of the N-BEATS-S loss function during training to further improve forecast stability while maintaining accuracy. Our contributions are threefold:
(i) we empirically demonstrate that some existing DLW methods can enhance forecast stability without sacrificing accuracy \textcolor{black}{and provide insights into why this might be the case};
(ii) we propose a novel variant of Random Weighting \citep{lin2022reasonable}, called Task-Aware Random Weighting (TARW), which allows for explicit prioritization of forecast accuracy;
\textcolor{black}{and} (iii) we experimentally validate TARW and other DLW methods, comparing their effectiveness in improving forecast stability (while maintaining accuracy) against N-BEATS-S with a tuned static $\lambda$.

We explain N-BEATS-S in detail in Section~\ref{SEC:LITERATURE}. 
Subsequently, we introduce the concept of DLW and provide an overview of existing algorithms in this domain.
In Section~\ref{SEC:METHODOLOGY}, we describe the various DLW strategies---including TARW---that will be used to optimize
N-BEATS-S in more detail.
Our experimental study is introduced in Section~\ref{SEC:EXPERIMENTAL_DESIGN},
with results presented and discussed in Section~\ref{SEC:RESULTS+DISCUSSION}.
Finally, Section~\ref{SEC:CONCLUSION} concludes the paper, \textcolor{black}{discusses its limitations,} 
and includes suggestions for future research.


\section{Related work} \label{SEC:LITERATURE}
In this section, we first explain the N-BEATS-S method and how it balances forecast accuracy and stability during optimization. Next, we introduce the concept of DLW.

\subsection{N-BEATS-S: stabilized N-BEATS forecasts}

\begin{figure}[h]
 \centering
	\includegraphics[height = 6 cm]{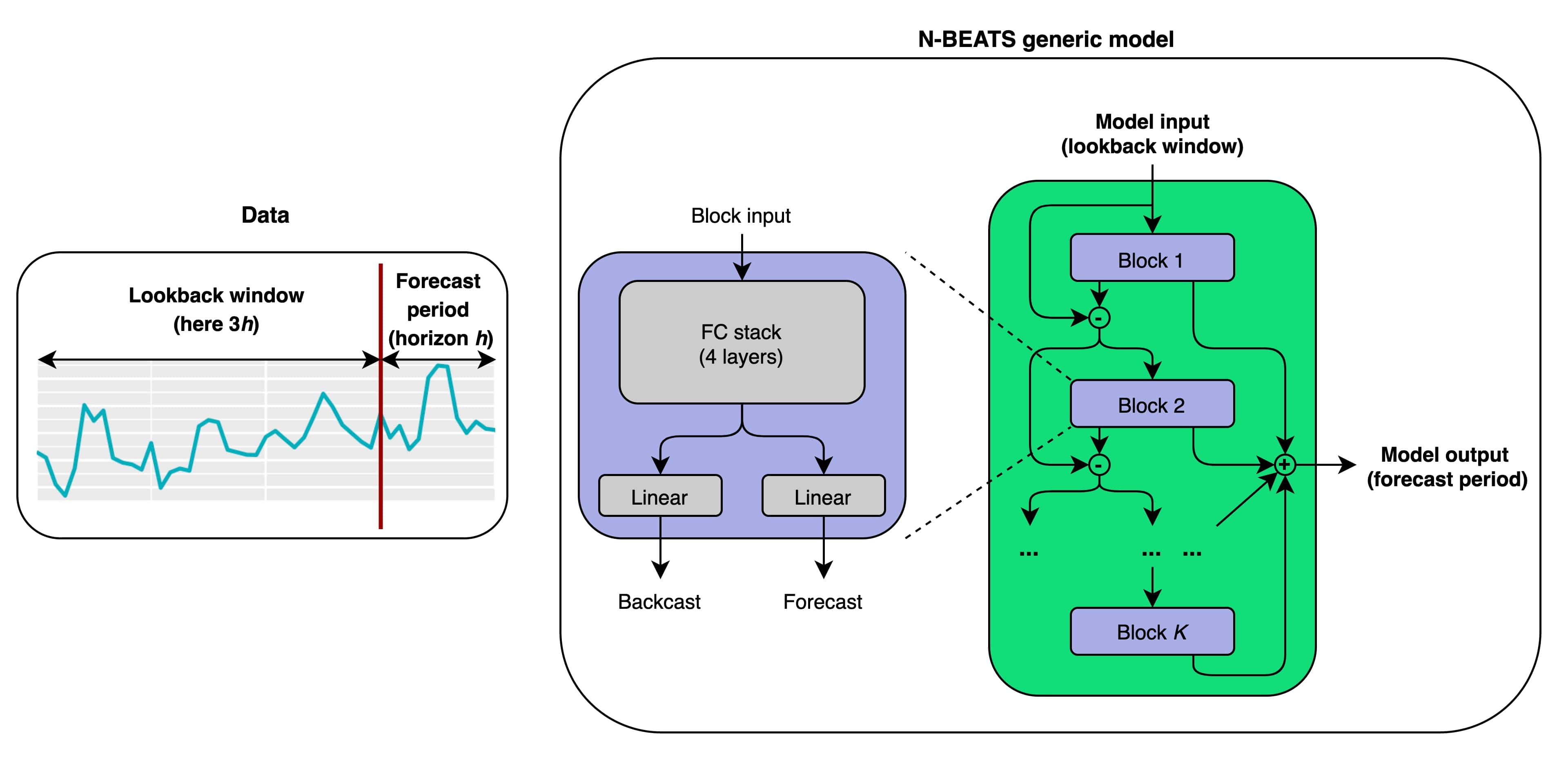}
	\caption{Generic N-BEATS architecture. Figure sourced from \citet{vanbelle2023}.
	\label{fig:N-BEATS}}
\end{figure}

The N-BEATS point forecasting method \citep{oreshkin2020nbeats} was the first pure deep learning approach to achieve state-of-the-art performance on the M3 and M4 data sets \citep{makridakis2000m3,makridakis2020m4}. As a global forecasting method, its parameters are optimized across different time series \citep{januschowski2020criteria}.
Its architecture, depicted in Figure~\ref{fig:N-BEATS}, consists of multiple 
processing blocks $k=1,\dots,K$, where $K$ is a hyperparameter,
organized using ``doubly residual stacking".
The basic building block of a generic\footnote{\citet{oreshkin2020nbeats} propose two N-BEATS configurations: a generic configuration and an interpretable configuration. In this work, we focus on the first one.} N-BEATS network consists of four fully connected layers, followed by two task-specific layers (one for each block output), all using ReLU activations.
Each block $k$ has an input vector $\vb{x}_k \in \mathbb{R}^T$ (the lookback window), 
where the lookback window length $T$ is a hyperparameter,
and produces two output vectors: a partial forecast $\hat{\vb{y}}_{k} \in \mathbb{R}^h$ (where $h$ is the forecast horizon) and a backcast $\hat{\vb{x}}_{k}$,
which is the block's best estimate of $\vb{x}_{k}$.
Backcasts are used to filter the input signal as it moves deeper into the network: the input for block $k+1$ is given by $\vb{x}_{k+1}=\vb{x}_k-\hat{\vb{x}}_k$, 
with $\vb{x}_1=\vb{x}$ containing the $T$ most recent observations at the forecasting origin. Partial forecasts are summed to produce the final forecasts $\hat{\vb{y}}=\sum_k\hat{\vb{y}}_k$ for the next $h$ observations $\vb{y} \in \mathbb{R}^h$.

To stabilize N-BEATS forecasts, \citet{vanbelle2023} propose N-BEATS-S,
an N-BEATS network optimized for both forecast accuracy and stability by relying on Equation~\eqref{EQ:L_COMBINED}.
More specifically,
given a training set of
$N$ 
input-output samples $\mathcal{D} = \{(\vb{x}^j_{T|t},\vb{y}_{h|t}^j)\}_{j=1}^{N}$,
where $\vb{x}^j_{T|t}$ contains the $T$ most recent observations at the forecasting origin $t$ ($y_{t-T+1}$ to $y_{t}$)\footnote{To simplify the notation, sample indexing is omitted for scalars.}
and $\vb{y}_{h|t}^j$ contains the next $h$ observations ($y_{t+1}$ to $y_{t+h}$),
they propose using an additional input-output pair 
$(\vb{x}^j_{T|t-1},\vb{y}_{h|t-1}^j)$
for each sample $j$ 
in order to quantify forecast instability via
$\mathcal{L}_\text{instability}(\hat{\vb{y}}_{h|t}^j,\hat{\vb{y}}_{h|t-1}^j)$, where $\hat{\vb{y}}_{h|t}^j$ and $\hat{\vb{y}}_{h|t-1}^j$ are the forecasts
for sample $j$ for the input-output pairs with forecasting origins $t$ ($\hat{y}_{t+1|t}$ to $\hat{y}_{t+h|t}$) and $t-1$ ($\hat{y}_{t|t-1}$ to $\hat{y}_{t+h-1|t-1}$), respectively.
The parameters $\bm{\theta}$ of an N-BEATS-S network $f(\vb{x};\bm{\theta})$ are then optimized as follows:
\begin{eqnarray}
    \bm{\theta}^* & = & \underset{\bm{\theta}}{\arg\min} \sum_{j} \bigg[
    \frac{1-\lambda}{2} \cdot \Big(\mathcal{L}_\text{error}(\mathbf{\hat{y}}_{h|t}^j,\mathbf{y}_{h|t}^j) + \mathcal{L}_\text{error}(\mathbf{\hat{y}}_{h|t-1}^j,\mathbf{y}_{h|t-1}^j)\Big) \nonumber \\
  & & \hspace{7cm} + \lambda \cdot \mathcal{L}_\text{instability}(\mathbf{\hat{y}}_{h|t}^j,\mathbf{\hat{y}}_{h|t-1}^j)
  \bigg], \\
 \mathbf{\hat{y}}_{h|t}^j & = & f(\mathbf{x}_{T|t}^j; \bm{\theta}), \\
 \mathbf{\hat{y}}_{h|t-1}^j & = & f(\mathbf{x}_{T|t-1}^j; \bm{\theta}).
\end{eqnarray}
For $\mathcal{L}_\text{error}$, they use the scale-independent 
root mean squared scaled error (RMSSE)
proposed by \citet{hyndman2006}:
\begin{equation} \label{EQ:RMSSE}
    \text{RMSSE}(\mathbf{\hat{y}}_{h|t}^j,\mathbf{y}_{h|t}^j) = \sqrt{ \frac{\frac{1}{h}\sum_{i=1}^{h}(y_{t+i}-\hat{y}_{t+i|t})^2}{\frac{1}{(T-1)}\sum_{i=1}^{T-1}(y_{t-i+1}-y_{t-i})^2}}.
\end{equation}
For $\mathcal{L}_\text{instability}$, 
\citet{vanbelle2023}
propose the root mean squared scaled change (RMSSC), 
defined similarly to RMSSE:
\begin{equation} \label{EQ:RMSSC}
    \text{RMSSC}(\mathbf{\hat{y}}_{h|t}^j,\mathbf{\hat{y}}_{h|t-1}^j) = \sqrt{ \frac{\frac{1}{(h-1)}\sum_{i=1}^{h-1}(\hat{y}_{t+i|t-1}-\hat{y}_{t+i|t})^2}{\frac{1}{(T-1)}\sum_{i=1}^{T-1}(y_{t-i+1}-y_{t-i})^2}},
\end{equation}
which quantifies the differences between
the model forecasts made at adjacent forecasting origins
\textcolor{black}{$t$ and $t-1$}
for the same overlapping time periods
\textcolor{black}{$t+1,\dots,t+h-1$}.

\subsection{Dynamic loss weighting (DLW)} \label{SEC:ADAPTIVE_LOSS_BALANCING}

In MTL, multiple tasks are trained in parallel. When these tasks are related, MTL can improve accuracy on one or more tasks compared to training a separate model for each task individually \citep{caruana1997multitask,ruder2017overview}. 
In MTL with deep learning, the primary objective is generally to learn shared representations for all tasks. 
These shared representations are typically followed by task-specific layers for each task.
Training these task-specific layers is straightforward: gradients are calculated with respect to the specific task loss, and stochastic gradient descent is used for optimization.
However, training the shared layers requires combining the task-specific losses, often by taking a weighted sum of these losses \citep{ruder2017overview}.

As explained in Section~\ref{SEC:INTRODUCTION}, 
the traditional approach to determining loss weights involves using grid search to find static values for these weights \citep{sener2018multi}.
However, this method may perform poorly due to several issues identified in the literature, 
such as different learning speeds
of the different tasks \citep{chen2018gradnorm} and conflicting gradients \citep{yu2020gradient}.
To address these issues, more recent works propose DLW algorithms.
These algorithms leverage the iterative nature of neural network optimization
by adjusting the loss weights dynamically during training, i.e., the loss weights can change throughout the process of optimizing the network parameters.
Some methods specifically target the issue of varying training rates,
which occurs because tasks may have different training dynamics. 
For instance, GradNorm \citep{chen2018gradnorm} and Dynamic Weight Average \citep{liu2019end} aim to balance the losses or gradients to ensure that different tasks learn at similar rates. Similarly, \citet{kendall2018multi} introduce Uncertainty Weighting (UW), a method based on quantifying task uncertainty to dynamically balance task-specific losses.
In contrast, \citet{yu2020gradient} address the problem of conflicting gradients by proposing a form of ``gradient surgery" to mitigate their influence.
They define two gradients to be conflicting if they point in opposite directions (i.e., have negative cosine similarity; see Section~\ref{SEC:METHODOLOGY} for a definition).
When the weighted gradient conflicts with one or more individual gradients, performance on the associated tasks will decrease.
Hence, since task gradients with larger magnitude can dominate the combined gradient, this may cause the optimizer to prioritize certain tasks over others.
More recently, \citet{lin2022reasonable} proposed Random Weighting (RW), a simple yet effective approach that involves randomly sampling loss weights in each training iteration.
This approach has been shown to achieve performance comparable to that of state-of-the-art techniques,
including those mentioned above,
on several multi-task computer vision and natural language processing problems. Consequently, the authors suggest that it should be considered a strong baseline.
\textcolor{black}{In contrast to the pragmatic RW approach, \citet{navon2022multi} propose a principled approach by adopting a game-theoretic perspective,
treating the gradient combination step as a cooperative bargaining game where tasks negotiate to reach an agreement on the update direction for the model parameters.}

A subfield within the MTL domain focuses on scenarios with one main task and one or more auxiliary tasks, with the goal of improving generalization performance on the main task.
Incorporating auxiliary tasks effectively enriches the learning problem with additional training data \citep{caruana1997multitask,ruder2017overview}. 
In this setting, dynamically adjusting the loss weights of the auxiliary tasks can be used to ensure that their gradients are used only if they benefit the main task's performance.
One intuitive approach is to use the auxiliary task gradient only if it has a positive cosine similarity 
(see Section~\ref{SEC:METHODOLOGY} for a definition)
with the main task gradient \citep{du2018adapting}.
\textcolor{black}{\citet{shamsian2023auxiliary} extend the method proposed by \citet{navon2022multi} by treating the gradient combination step as a generalized bargaining game with asymmetric task bargaining power.}
Other approaches determine the loss weights for auxiliary tasks by using a stable main task metric calculated over multiple minibatches \citep{verboven2022hydalearn} or by employing a holdout main task metric \citep{gregoire2024sample}.

Recall from Section~\ref{SEC:INTRODUCTION} that our goal is to explore the potential of dynamically weighting the two components of the N-BEATS-S loss function during training to further improve forecast stability, compared to using static loss weights, without compromising accuracy.
Given this constraint, we can approach this problem from an auxiliary task learning perspective, treating forecast accuracy as the main task and forecast stability as the auxiliary task. 
However, unlike traditional auxiliary task learning settings---where the final performance on auxiliary tasks is generally of lesser importance---we are explicitly interested in improving forecast stability (while either improving or at least maintaining forecast accuracy).
Therefore, the problem we address in this paper can be situated between the auxiliary task learning setting and the general MTL setting.






\section{Optimizing N-BEATS-S with DLW methods}
\label{SEC:METHODOLOGY}

\textcolor{black}{Training N-BEATS-S using a DLW method leverages the iterative nature of neural network optimization to dynamically adjust the impact of forecast accuracy and forecast stability on the network parameters during training. This is achieved by using a different $\lambda_i$ for each learning iteration $i = 1,2,\dots$, as conceptually summarized in Pseudocode~\ref{alg:pseudo}.}
DLW methods differ in how they dynamically calculate $\lambda_i$ (line 4).

\begin{algorithm}
	\caption{Training N-BEATS-S with DLW\protect\footnotemark.}
        \label{alg:pseudo}
        \nonumber
	\begin{algorithmic}[1]
		\For {training iteration $i=1,2,\ldots$ with minibatch $\mathcal{D}_i$ and learning rate $\eta$}
                \State Compute loss terms $\mathcal{L}_\text{error}^i(\mathcal{D}_i;\bm{\theta})$ and $\mathcal{L}_\text{instability}^i(\mathcal{D}_i;\bm{\theta})$;
                \State Compute gradients $\vb{g}_\text{error}^i = \nabla_{\bm{\theta}}\mathcal{L}_\text{error}^i$ and $\vb{g}_\text{instability}^i = \nabla_{\bm{\theta}}\mathcal{L}_\text{instability}^i$;
                \State \colorbox{lightgray}{Compute $\lambda_i$};
                \State $\bm{\theta} \leftarrow \bm{\theta} - \eta \bigl((1-\lambda_i) \vb{g}_\text{error}^i + \lambda_i \vb{g}_\text{instability}^i\bigr)$;
            \EndFor
	\end{algorithmic} 
\end{algorithm}
\footnotetext{In an MTL network with both shared and task-specific layers, and therefore both shared and task-specific parameters, a distinction can be made between gradient weighting and loss weighting \citep[see, e.g.,][]{lin2022reasonable}. Gradient weighting methods use the weights only for updating the shared parameters, while the task-specific parameters are updated using the unweighted task-specific gradient.
In contrast, loss weighting methods use the weights for updating all parameters (which affects the learning rate for the task-specific parameters).
Since N-BEATS-S only has shared parameters, we do not explicitly differentiate between these two approaches in this paper.}

We will investigate the performance of the following existing DLW methods:
\begin{itemize}
    \item GradNorm \citep{chen2018gradnorm}: $\lambda_i$ is updated to bring the gradient norms of the different tasks closer together to balance their training rates. A hyperparameter $\alpha$ controls the strength of this balancing, with higher values enforcing stronger equalization of training rates. An initial value for $\lambda_0$ needs to be set to initialize the algorithm.
    \item Uncertainty Weighting (UW) \citep{kendall2018multi}: 
    $\lambda_i$ is updated
    based on the learned relative homoscedastic uncertainties of the different tasks.
    A task's homoscedastic uncertainty
    reflects the uncertainty inherent to the task.
    It is the aleatoric uncertainty (inherent randomness in the data) that stays constant for all input data but varies between different tasks.
    If a task's relative homoscedastic uncertainty increases, its weight is decreased, and vice versa.
    \textcolor{black}{An initial value for $\lambda_0$ needs to be set to initialize the algorithm.}
    \item \textcolor{black}{NashMTL \citep{navon2022multi}: $\lambda_i$ is determined by framing the gradient combination step as a cooperative bargaining game, where tasks negotiate to reach an agreement on the update direction for the model parameters. Under certain assumptions, this bargaining problem has a unique solution known as the Nash bargaining solution. 
    This solution, which is both Pareto optimal and proportionally fair (i.e., treating all tasks as equally important), is approximated to compute the loss weights in each iteration.}
    \item Random Weighting (RW) \citep{lin2022reasonable}: $\lambda_i$ is randomly sampled from a standard uniform distribution, $U(0,1)$.
    \item Gradient Cosine Similarity (GCosSim) \citep{du2018adapting}: $\lambda_i$ is either 0 or 0.5, depending on the cosine similarity between
    $\vb{g}_\text{error}^i$ (considered the main task's gradient)
    and
    $\vb{g}_\text{instability}^i$ (considered the auxiliary task's gradient), given by $\frac{\vb{g}_\text{error}^i \cdot \vb{g}_\text{instability}^i}{\| \vb{g}_\text{error}^i \| \| \vb{g}_\text{instability}^i \|} \in [-1,1]$. 
    If the cosine similarity is negative, 
    $\lambda_i=0$, and forecast instability is ignored.
    If the cosine similarity is positive, $\lambda_i=0.5$, and the gradients of both tasks are summed to update the model parameters.
    \item Weighted GCosSim \citep{du2018adapting}: This method is similar to GCosSim but accounts for the degree of similarity between $\vb{g}_\text{error}^i$ and $\vb{g}_\text{instability}^i$ (i.e., the degree to which these two vectors point in the same direction).
    If the cosine similarity is positive but less than one, indicating only partial alignment, $\lambda_i$ equals half of the cosine similarity.
    \item \textcolor{black}{AuxiNash \citep{shamsian2023auxiliary}:
    This method generalizes NashMTL by allowing for asymmetric task bargaining power to account for varying preferences for forecast stability ($p_i$) and forecast accuracy ($1-p_i$).
    The preference parameter $p_i$ is dynamically learned during training by optimizing for maximal performance on the main task (i.e., forecast accuracy), approximated using a randomly sampled separate training batch. Specifically, $p_i$ is updated every hyperstep ($hyp$, i.e., the preference update rate) with a distinct learning rate $\eta_p$ and an initial value $p_{init}$.}
\end{itemize}
Keeping in mind our goal of further improving forecast stability without compromising accuracy by optimizing N-BEATS-S with a DLW method instead of static loss weights, and considering that this problem can be situated between the auxiliary task learning setting and the general MTL setting (see Section~\ref{SEC:ADAPTIVE_LOSS_BALANCING}), we propose a variant of RW to better fit this context:
\begin{itemize}
    \item Task-Aware Random Weighting (TARW):
    $\lambda_i$ is randomly sampled from a uniform distribution, $U(0,\kappa)$, where $\kappa \in (0, 1]$ is a tunable hyperparameter that caps the maximum value of the uniform distribution, preventing excessively high weights from being assigned to forecast instability. Note that $\kappa$ remains the same over all iterations.
\end{itemize}
Where RW can be seen as the stochastic version of equal weighting (where each task is assigned the same weight) \citep{lin2022reasonable},
TARW can be considered the stochastic version of static loss weight tuning. Instead of treating $\lambda$ as a static hyperparameter, we obtain dynamic loss weights ($1-\lambda_i$) and $\lambda_i$ by tuning the hyperparameter $\kappa$, which characterizes the uniform distribution from which the weights are sampled.

\section{Experimental design} \label{SEC:EXPERIMENTAL_DESIGN}

In this section, we provide a detailed description of the experimental design.
We begin by describing the data sets and the evaluation scheme used.
Next, we present an overview of the forecasting methods included for comparison in our study and briefly outline the adopted training methodology. Finally, we explain how the hyperparameter values were obtained.
The code to reproduce the experiments is available online at \url{https://anonymous.4open.science/r/Dynamic-N-BEATS-S-223B/}.

\subsection{Data sets}

We use the monthly time series from the M3 \citep{makridakis2000m3} and M4 \citep{makridakis2020m4} data sets, with summary statistics presented in Table~\ref{TAB:DATA_SETS}. All series have positive observed values at every time step.

\begin{table}[!htbp]
	\centering
	\begin{tabular}{lcc}
            \hline
		& \multicolumn{1}{l}{{M3 monthly}} & \multicolumn{1}{l}{{M4 monthly}} \\
		\hline
		\multicolumn{1}{l}{No. of series} & 1,428 & 48,000 \\
		\multicolumn{1}{l}{Min. length} & 66 & 60 \\
		\multicolumn{1}{l}{Max. length} & 144 & 2812 \\
		\multicolumn{1}{l}{Mean length} & 117.3 & 234.3 \\
		\multicolumn{1}{l}{Std. dev. length} & 28.5 & 137.4 \\
		\hline
	\end{tabular}
	\captionsetup{justification=centering}
	\caption{Summary statistics of the data sets used.}
	\label{TAB:DATA_SETS}	
\end{table}

In both the original M3 and M4 competitions, participants were tasked with generating one- to 18-month-ahead out-of-sample forecasts from a single forecasting origin. Specifically, the test set included the last 18 data points of each time series \citep{makridakis2000m3,makridakis2020m4}. We follow this setup and use the same test set to report performance metrics.

\subsection{Evaluation scheme}

We adopt the evaluation scheme used by \citet{vanbelle2023} to be able to evaluate forecast stability in addition to forecast accuracy: for both data sets, a rolling forecasting origin evaluation \citep{tashman2000} is performed for each time series, in which one- to six-month-ahead forecasts from 13 consecutive forecasting origins are evaluated (so as to use the full test set, i.e., the last 18 observations of each time series).
The results presented in Section~\ref{SEC:RESULTS+DISCUSSION} are averaged across 
(different pairs of)
forecasting origins\footnote{Due to the use of different evaluation schemes, our results are not directly comparable to those reported in the literature for the M3 and M4 data sets.} and then averaged again across all time series.

\textcolor{black}{To evaluate forecast accuracy and stability, we use the RMSSE and RMSSC as defined in Equations~\eqref{EQ:RMSSE} and \eqref{EQ:RMSSC}, with $T=t$ (i.e., all historical values available up to the forecasting origin $t$ are used to calculate the denominator in these equations).
Additionally, \ref{sec:results_appendix} reports results in terms of the scale-independent symmetric mean absolute percentage error (sMAPE), which was used in the M3 and M4 competitions to evaluate forecast accuracy \citep{makridakis2000m3,makridakis2020m4}, and the scale-independent symmetric mean absolute percentage change (sMAPC), 
which was proposed by \citet{vanbelle2023} to evaluate forecast stability and is defined similarly to sMAPE.}

\subsection{Forecasting methods} \label{SEC:FORECASTING_METHODS}

Alongside the results for N-BEATS-S optimized using the various DLW methods discussed in Section~\ref{SEC:METHODOLOGY}, we also report results for the following methods:
\begin{itemize}
    \item N-BEATS: A standard N-BEATS model as described in \citet{oreshkin2020nbeats}. To fairly compare N-BEATS and N-BEATS-S forecasts, we also use the additional input-output pairs for the N-BEATS model (even though they are strictly unnecessary), with $\lambda$ set to zero to ignore the instability loss term.
    \item N-BEATS-S: N-BEATS-S with a static value for $\lambda$ \citep{vanbelle2023}.
    \item ETS: An automatically selected exponential smoothing model using the \texttt{ets()} function from the \texttt{forecast} R package \citep{hyndman2008ETS,hyndman2008forecastmanual}.
    \item ARIMA: An automatically selected ARIMA model using the \texttt{auto.arima()} function from the \texttt{forecast} R package \citep{hyndman2008automatic}.
    \item THETA: The method proposed by \citet{assimakopoulos2000theta}, which won the M3 competition \citep{makridakis2000m3}. Forecasts are obtained using the \texttt{thetaf()} function from the \texttt{forecast} R package. 
\end{itemize}
As in \citet{vanbelle2023}, for each N-BEATS(-S) variant, we run the network with its specific set of hyperparameter values five times\footnote{\citet{oreshkin2020nbeats} presented results using ensembles of 180 N-BEATS networks.},
each with a different initialization. 
The medians of the forecasts from these runs are then used as the final forecasts.

\subsection{Training methodology for N-BEATS(-S) networks} \label{SEC:TRAININGMETHOD}


\textcolor{black}{Following \citet{vanbelle2023}, we use the scale-independent 
RMSSE and RMSSC for the forecast error and instability loss terms, respectively, thereby eliminating the need for data preprocessing.}
All networks are implemented in PyTorch \citep{NEURIPS2019_9015}, and we use the Adam optimizer with \textcolor{black}{its} default settings \citep{kingma2014} 
and initial learning rates specified in Table~\ref{TAB:HYPERPARAMETERS}
to optimize the \textcolor{black}{network} parameters.
To construct training batches, \textcolor{black}{we also follow the approach outlined in \citet{vanbelle2023}. First, time series are sampled uniformly at random.
Next, to obtain an input-output sample for each selected time series, a time step is sampled uniformly at random from the forecasting origin range.}
This range comprises the most recent observations that do not result in missing values when creating a sample, and its size is a hyperparameter.

\subsection{Hyperparameters for N-BEATS(-S) networks} \label{SEC:HYPERPARS}

Table~\ref{TAB:HYPERPARAMETERS} provides an overview of the hyperparameter values used for training all N-BEATS(-S) networks,
which are the tuned 
values reported in \citet{vanbelle2023}. 
\textcolor{black}{For all N-BEATS(-S) variants in this study, only the additional hyperparameters, including the learning rate and the number of learning iterations, are tuned, while the other hyperparameters are 
set to the values listed in Table~\ref{TAB:HYPERPARAMETERS}.
Table~\ref{tab:combined_hyperparameters} provides an overview of the selected values for the additional hyperparameters.}

\begin{table}[!htbp]
	\centering
	\begin{tabular}{lccc}
            \hline
		& M3 monthly & \phantom{a} & M4 monthly \\
		\hline
		No. of blocks \textcolor{black}{$K$} & 20 & & 20 \\
		Hidden layer width & 256 & & 256 \\
		Batch size & 512 & & 512 \\
		Lookback window length $T$ & 6$h$ & & 4$h$ \\
		Forecasting origin range & 20$h$ & & 10$h$ \\
		\hline
	\end{tabular}
	\caption{\textcolor{black}{Hyperparameters for N-BEATS(-S) networks from \citet{vanbelle2023}.}}
	\label{TAB:HYPERPARAMETERS}
\end{table}

\begin{table}[!htbp]
\centering
\resizebox{\textwidth}{!}{%
\begin{tabular}{lccccccc}
\hline
& \multicolumn{3}{c}{M3 monthly} & \phantom{a} & \multicolumn{3}{c}{M4 monthly} \\
\cline{2-4} \cline{6-8}
& Iterations & Learning rate & Other & & Iterations & Learning rate & Other \\
\hline
N-BEATS & 8,000 & 1e-5 &  & & 14,415 & 1e-3 & \\
\hline
\textcolor{black}{N-BEATS-S low}\footnotemark & 8,000 & 1e-5 & $\lambda=0.025$& & 14,415 & 1e-3 & $\lambda=0.025$\\
\textcolor{black}{N-BEATS-S high} & 8,000 & 1e-5 & $\lambda=0.075$ & & 14,415 & 1e-3 &$\lambda=0.275$ \\
\hdashline
GradNorm & 8,000 & 1e-5 & $\alpha=0$; $\lambda_0=0.05$ & & \textcolor{black}{18,600} & 1e-3 & \textcolor{black}{$\alpha=1$}; $\lambda_0=0.05$ \\
UW & 10,000 & 1e-5 & $\lambda_0=0.5$ & & 14,415 & 1e-5 & $\lambda_0=0.5$\\
\textcolor{black}{NashMTL} & 10,000 & 1e-5 & & & 14,415 & 5e-4 & \\
RW & 9,000 & 1e-5 & & & 18,600 & 1e-4 & \\
\hdashline
GCosSim & 8,000 & 1e-5 & & & 14,415 & 1e-3 & \\
Weighted GCosSim & 8,000 & 1e-5 & & & 14,415 & 5e-4 & \\
\textcolor{black}{AuxiNash} & 10,000 & 1e-5 & $hyp = 10, \eta_p = 0.05, p_{init} = 0.1$ & & 18,600 & 5e-4 & $hyp = 10, \eta_p = 0.01, p_{init} = 0.05$ \\
\textit{\textcolor{black}{TARW low}} & 10,000 & 1e-5 & $\kappa = 0.125$ & & 23,808 & 1e-4 & $\kappa = 0.125$ \\
\textit{\textcolor{black}{TARW high}} & 10,000 & 1e-5 & $\kappa = 0.20$ & & 23,808 & 1e-4 & $\kappa = 0.20$ \\
\hline
\end{tabular}%
}
\captionsetup{justification=centering}
\caption{\textcolor{black}{Additional hyperparameters for 
N-BEATS(-S)
networks.}}
\label{tab:combined_hyperparameters}
\end{table}

\textcolor{black}{The learning rates are tuned} because some methods exhibit unstable training curves with the values reported in
\textcolor{black}{\cite{vanbelle2023}}.
The number of iterations is also adjusted for two reasons: (i) a lower learning rate may require more iterations for the model to converge, and (ii) more advanced DLW methods might need additional iterations or data to achieve convergence.
These hyperparameters are optimized using grid search and selected based on the minimum validation \textcolor{black}{RMSSE} (for converged validation losses) from a single run.
The validation \textcolor{black}{RMSSE} is calculated on a holdout set
comprising the 18 observations immediately preceding the first data point in the test set for each time series.
The rolling origin evaluation procedure used for the test set results is also applied to obtain validation set results.

\footnotetext{\textcolor{black}{This corresponds to the tuning strategy used in \citet{vanbelle2023}. Note, however, that the selected $\lambda$ values differ from those reported in \citet{vanbelle2023} because they were selected based on validation RMSSE instead of validation sMAPE.}}

\textcolor{black}{For N-BEATS-S with static loss weights and TARW, the hyperparameters $\lambda$ and $\kappa$ are crucial as they directly control the extent to which the forecast instability component affects the optimization problem and therefore the model outputs.
In line with \citet{vanbelle2023}, 
who report that there is a wide range of $\lambda$ values for N-BEATS-S that lead to improvements in both forecast accuracy and stability (compared to $\lambda=0$), 
we observe a similar pattern for $\kappa$ values in TARW (see Section~\ref{sec:TARW}).
Following \citet{vanbelle2024}, 
who propose an N-BEATS variant to stabilize Gaussian probabilistic forecasts that requires tuning a hyperparameter analogous to $\lambda$, 
we report results for both conservatively and more aggressively selected values for $\lambda$ and $\kappa$.
N-BEATS-S low and TARW low are based on conservative values obtained by minimizing validation RMSSE,
while N-BEATS-S high and TARW high use the largest $\lambda$ and $\kappa$ values for which no clear deterioration in validation RMSSE is observed compared to $\lambda=0$ and $\kappa=0$, respectively.
For these crucial hyperparameters, validation RMSSE (and RMSSC) is smoothed by averaging the RMSSE (and RMSSC) results from two runs with different random initializations.
N-BEATS-S low and TARW low are expected to yield smaller improvements in stability while minimizing the risk of accuracy loss, whereas N-BEATS-S high and TARW high are expected to achieve larger improvements in stability, albeit with a higher risk of (slight) decreases in accuracy.}


After the hyperparameters are tuned, the networks are trained on the entire training set, including the validation set. These trained networks are then used to generate forecasts for the test set, which are used to report performance metrics.


\section{Results and discussion} \label{SEC:RESULTS+DISCUSSION}

In this section, we present and discuss the results of the experiments conducted to evaluate the impact of using a DLW method to train N-BEATS-S,
focusing on its ability to further improve the stability of N-BEATS-S forecasts 
without sacrificing forecast accuracy.

\subsection{Results} \label{SEC:RESULTS}

Table~\ref{tab:results} summarizes the test set results for the M3 and M4 monthly data sets. 
All DLW algorithms outperform N-BEATS-S \textcolor{black}{low} on both data sets in terms of \textcolor{black}{RMSSC}, which measures forecast stability.
However, our goal is to improve forecast stability while either improving or at least maintaining forecast accuracy.
For the M3 data set, TARW \textcolor{black}{high} achieves the highest accuracy among the DLW variants, outperforming both N-BEATS and N-BEATS-S. \textcolor{black}{AuxiNash, TARW low, and Weighted Gcossim} also have RMSSE values close to those of N-BEATS and N-BEATS-S, while UW, RW, and GCosSim fail to maintain accuracy.
For the M4 data set, the results are similar, with TARW \textcolor{black}{(both high and low)} being the best-performing DLW variant \textcolor{black}{in terms of accuracy}, achieving \textcolor{black}{almost the same RMSSE as N-BEATS-S low. TARW is followed closely by both AuxiNash and Weighted GcosSim, while N-BEATS-S high performs worse than N-BEATS due to its use of a relatively large $\lambda$ value, which also results in a substantial boost in stability (see Section~\ref{SEC:HYPERPARS}).}
Although UW, RW, and GCosSim perform best in terms of stability on both data sets, they do so at the cost of considerable accuracy, making them less suitable for the problem addressed in this study.
Finally, note that 
all DLW variants outperform the traditional time series forecasting methods (ETS, ARIMA, and THETA) in terms of stability.
However, in terms of accuracy, the local ETS and \textcolor{black}{ARIMA} models outperform the deep learning models on the M3 data set, 
\textcolor{black}{and ARIMA remains competitive with the best-performing DLW variants on the larger M4 data set.}

\begin{table}[!htbp]
\centering
{\color{black}
\begin{tabular}{lccccc}
\hline
& \multicolumn{2}{c}{{M3 monthly}} & & \multicolumn{2}{c}{{M4 monthly}} \\
\cline{2-3} \cline{5-6}
& {RMSSE} & {RMSSC} & & {RMSSE} & {RMSSC} \\
\hline
\multicolumn{1}{l}{N-BEATS} & 1.088 & 0.393 & & 1.266 & 0.556 \\
\hline
\multicolumn{1}{l}{N-BEATS-S low} & 1.089 &0.382 & & \textbf{1.256} & 0.537 \\
\multicolumn{1}{l}{N-BEATS-S high} & 1.088 &0.352 & & {1.274} & 0.364 \\
\hdashline
\multicolumn{1}{l}{GradNorm} & 1.122 &  0.226 & & 1.299 & 0.324 \\
\multicolumn{1}{l}{UW} &1.172 &\textbf{0.145} & &1.526 &\textbf{0.146} \\
\multicolumn{1}{l}{NashMTL} & 1.124 & 0.251 & & 1.278& 0.339 \\
\multicolumn{1}{l}{RW} & 1.174 & \underline{0.165} & & 1.354 & \underline{0.224} \\
\hdashline
\multicolumn{1}{l}{GCosSim} & 1.165 & {0.157} & & 1.350 & {0.234} \\
\multicolumn{1}{l}{Weighted GCosSim} &  1.110 &  0.271 & & 1.265 & 0.384 \\
\multicolumn{1}{l}{AuxiNash} & 1.101 & 0.297 & & 1.260 & 0.410 \\
\multicolumn{1}{l}{\textit{TARW low}} & 1.087 & 0.363 & & $\underline{1.257}$ & 0.499 \\
\multicolumn{1}{l}{\textit{TARW high}} & 1.082 & 0.349 & & $\underline{1.257}$ & 0.460\\
\hline
\multicolumn{1}{l}{ETS} & \underline{1.048} & 0.411 & & 1.322 & 0.594 \\
\multicolumn{1}{l}{ARIMA} & \textbf{1.044} & 0.419 & & 1.271 & 0.571 \\
\multicolumn{1}{l}{THETA} & {1.094} & 0.366 & & 1.406 & 0.526 \\
\hline
\end{tabular}
}
\captionsetup{justification=centering,font={color=black}}
\caption{Forecast accuracy and stability performance on the
test sets. 
Lower is better.\\
The minimum value per column is highlighted in bold.}
\label{tab:results}
\end{table}

\begin{figure}[!htbp]
\centering
\begin{subfigure}[b]{0.49\textwidth}
\centering
\includegraphics[width=1\linewidth]{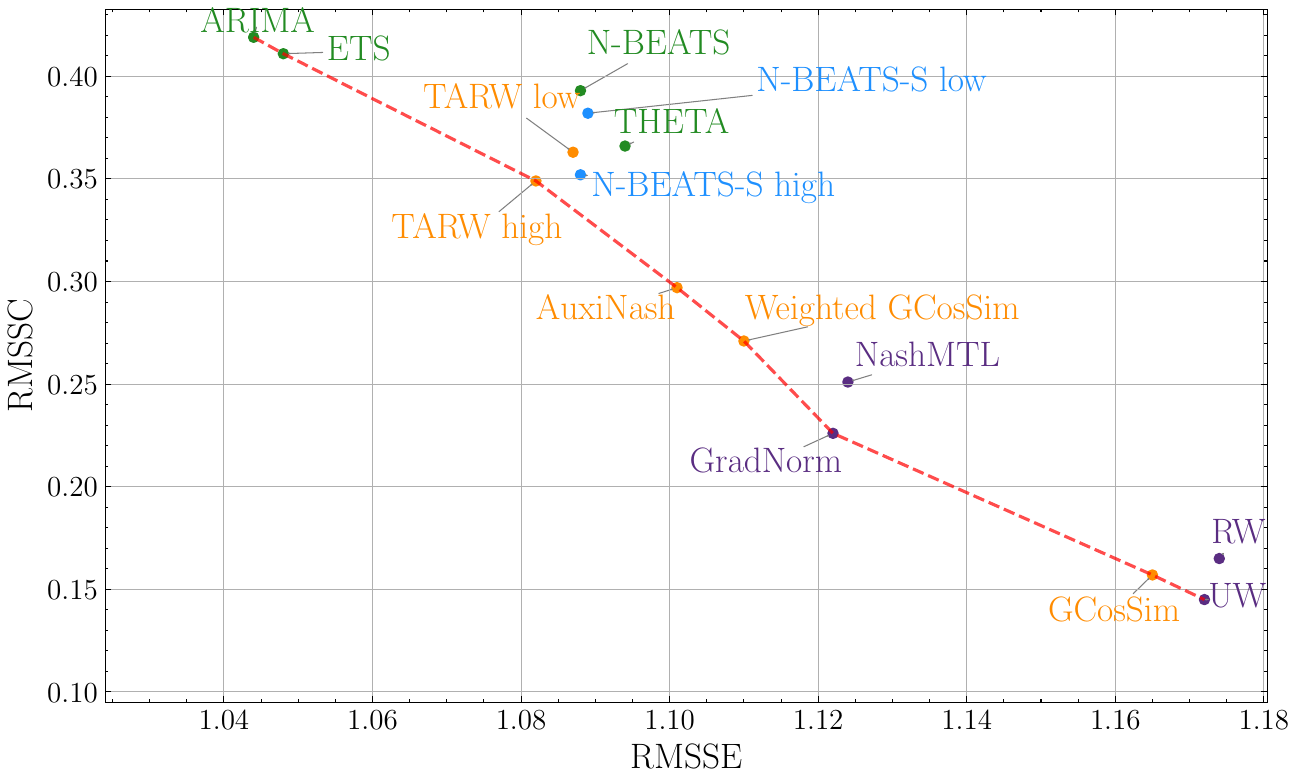}
\caption{M3 monthly} 
\label{fig:pareto M3}
\end{subfigure}
\begin{subfigure}[b]{0.49\textwidth}
\centering
\includegraphics[width=1\linewidth]{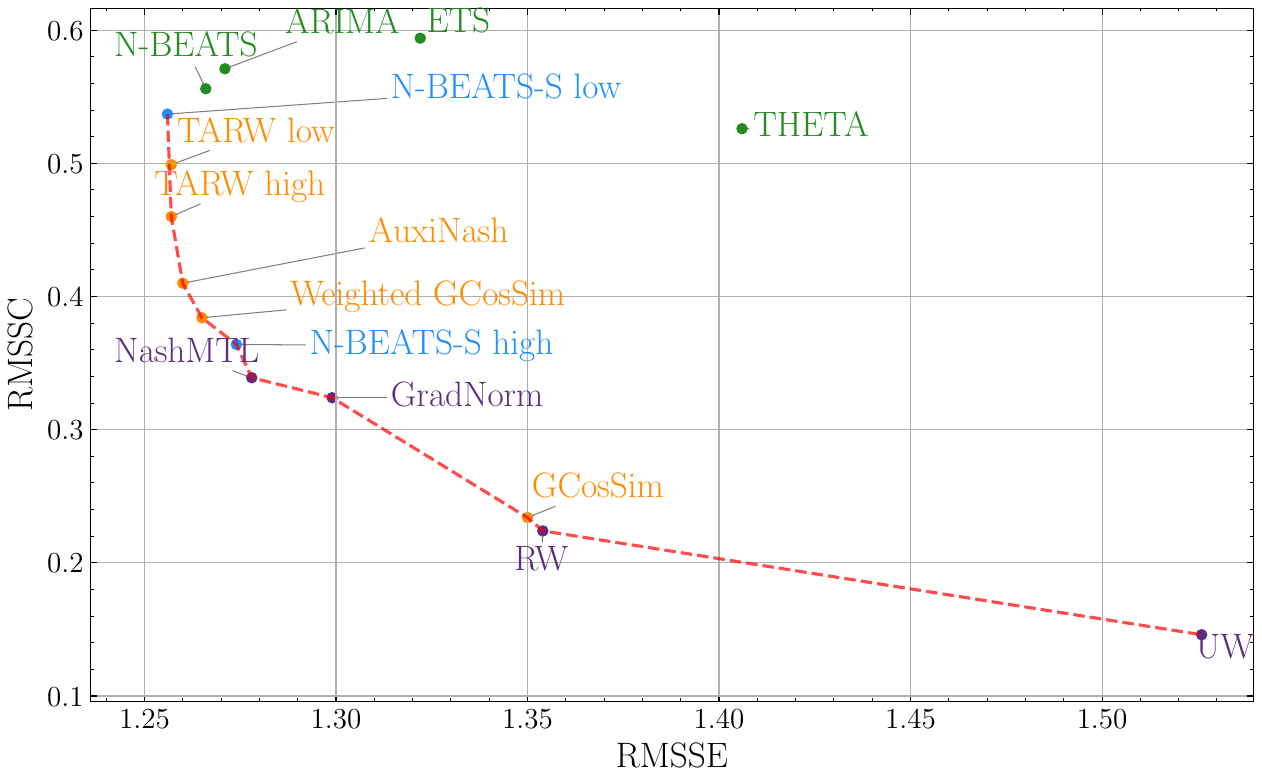}
\caption{M4 monthly} 
\label{fig:pareto M4}
\end{subfigure}
\captionsetup{justification=centering}
\caption{\textcolor{black}{Pareto frontiers for the M3 and M4 data sets.}} 
\label{fig:Pareto plots}
\end{figure}

\textcolor{black}{The results from Table~\ref{tab:results} are visually represented in Figure \ref{fig:Pareto plots}, facilitating the construction of Pareto frontiers.
A method is Pareto efficient if no other method achieves better accuracy for the same (or better) stability, or vice versa.
All DLW variants, except for NashMTL and RW on the M3 data set, are Pareto efficient but take different positions on the Pareto frontiers, reflecting varying accuracy-stability trade-offs.
An additional observation is that, while N-BEATS is Pareto inefficient on both the M3 and M4 data sets, N-BEATS-S low and N-BEATS-S high are Pareto inefficient only on the M3 data set.}

\begin{figure}[!htbp]
\centering
\begin{subfigure}[b]{0.49\textwidth}
\centering
\includegraphics[width=1\linewidth]{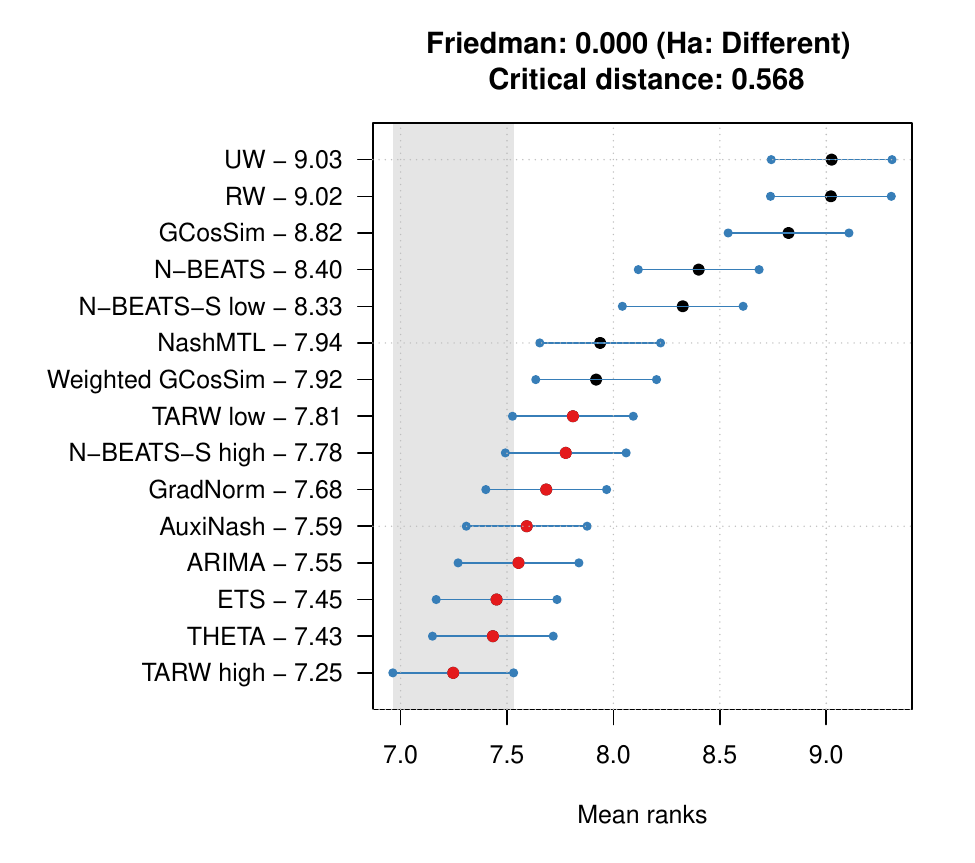}
\caption{RMSSE ranking} 
\label{fig:MCB M3 acc}
\end{subfigure}
\begin{subfigure}[b]{0.49\textwidth}
\centering
\includegraphics[width=1\linewidth]{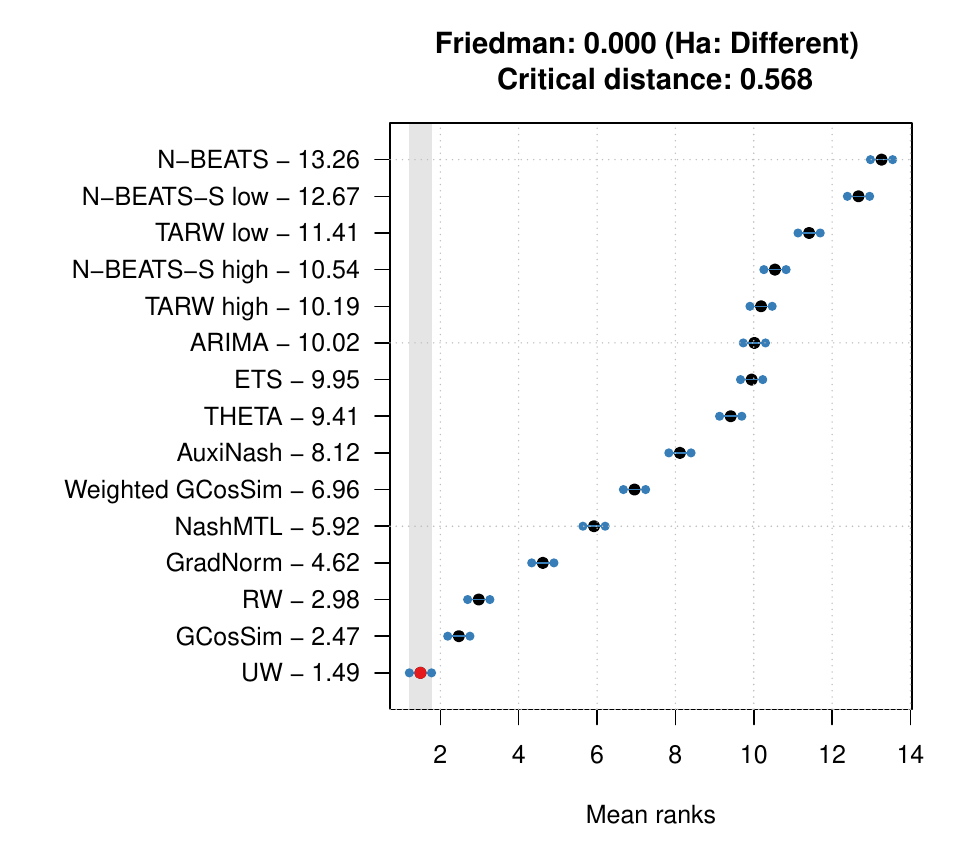} 
\caption{RMSSC ranking}
\label{fig:MCB M3 stab}
\end{subfigure}
\captionsetup{justification=centering}
\caption{\textcolor{black}{MCB results for the M3 monthly data set. 
Lower is better. 
If two intervals overlap, there is no statistically significant difference between the corresponding methods.}
\label{fig:MCB M3}}
\end{figure}

\begin{figure}[!htbp]
\centering
\begin{subfigure}[b]{0.49\textwidth}
\centering
\includegraphics[width=1\linewidth]{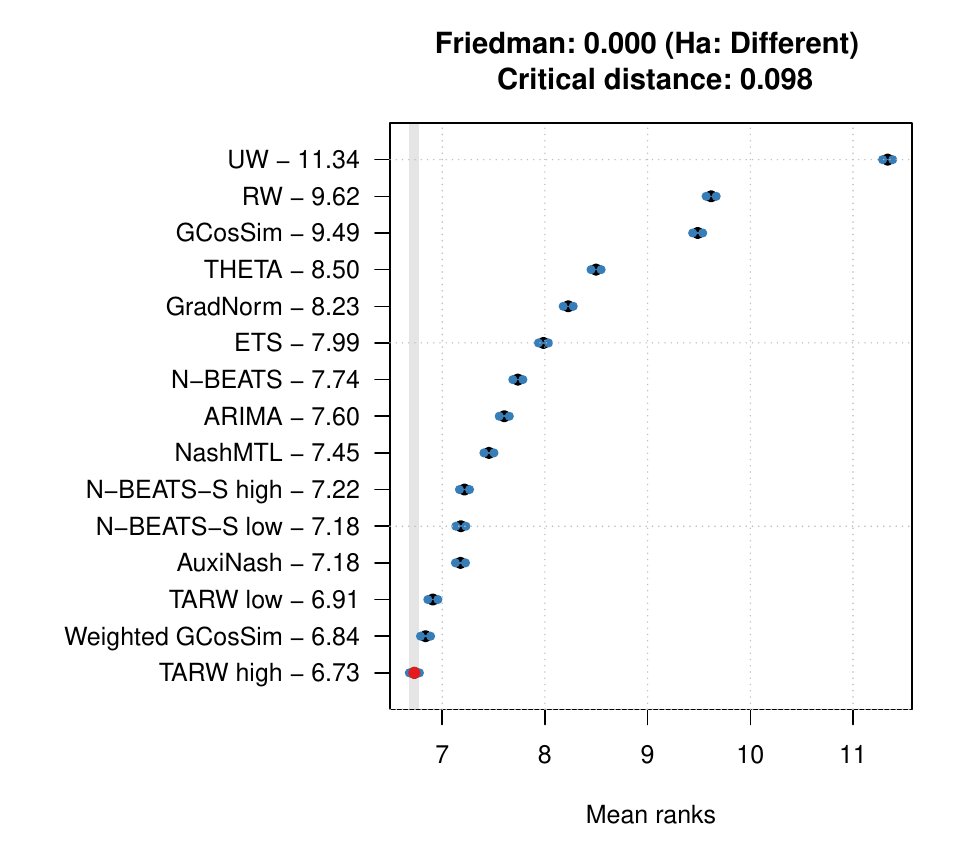}
\caption{RMSSE ranking} 
\label{fig:MCB M4 acc}
\end{subfigure}
\begin{subfigure}[b]{0.49\textwidth}
\centering
\includegraphics[width=1\linewidth]{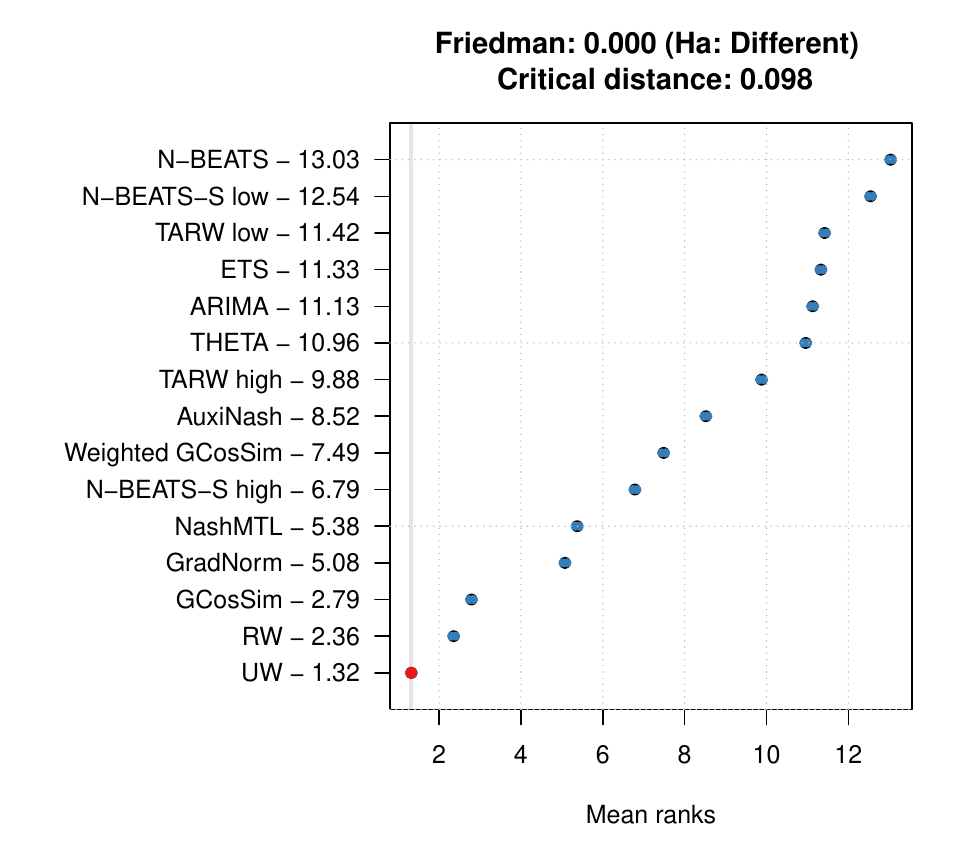} 
\caption{RMSSC ranking}
\label{fig:MCB M4 stab}
\end{subfigure}
\captionsetup{justification=centering}
\caption{\textcolor{black}{MCB results for the M4 monthly data set. 
Lower is better.
If two intervals overlap, there is no statistically significant difference between the corresponding methods.}
\label{fig:MCB M4}}
\end{figure}

To determine whether the reported differences in \textcolor{black}{RMSSE and RMSSC} are statistically significant,
we also present results from multiple comparisons with the best (MCB) tests \citep{koning2005m3}.
The MCB test calculates the average rank of each method 
across all time series in a data set 
based on a specified performance metric
and constructs an interval around this average. If the intervals of two methods do not overlap, the difference between these methods is statistically significant.
The results of the MCB tests are shown in Figure~\ref{fig:MCB M3} (M3 monthly) and Figure~\ref{fig:MCB M4} (M4 monthly), with the interval of the best method highlighted by the grey-shaded area. Additionally, all methods can be compared by examining the overlaps or gaps between their intervals.

The MCB results for the M3 data set indicate that \textcolor{black}{TARW high and the local forecasting methods} generate the most accurate forecasts in terms of average RMSSE rank,
with a large group of methods,
\textcolor{black}{including N-BEATS-S high, and the DLW variants AuxiNash, GradNorm, and TARW low,}
resulting in forecasts with similar average accuracy rank.
\textcolor{black}{Moreover, the results confirm that both AuxiNash and GradNorm also generate statistically significant improvements in stability compared to the best-performing N-BEATS-S variant with a static $\lambda$ value (and the traditional time series methods).}
For the M4 data set, \textcolor{black}{TARW high} produces the most accurate forecasts in terms of average \textcolor{black}{RMSSE} rank,
followed by Weighted GCosSim and \textcolor{black}{TARW low.}
\textcolor{black}{These three DLW variants  also significantly outperform N-BEATS-S low} in terms of forecast stability.

\textcolor{black}{By considering both the results in terms of average RMSSE and RMSSC, 
as well as in terms of the corresponding average ranks,}
we can conclude that 
Weighted GCosSim,
\textcolor{black}{AuxiNash},
and TARW further improve forecast stability compared to N-BEATS-S without considerably compromising forecast accuracy.
Among these three DLW variants, TARW performs best in terms of accuracy but yields the smallest improvements in stability.

\subsection{Discussion} \label{SEC:DISCUSSION}

In this section, we further investigate
GradNorm, Weighted GCosSim, \textcolor{black}{AuxiNash}, 
and TARW to gain a deeper understanding of \textcolor{black}{how} the underlying mechanisms of these DLW methods
\textcolor{black}{lead to further improvements in forecast stability as compared to N-BEATS-S with static loss weights.}

\subsubsection{GradNorm} \label{SEC:DISCUSSION_GRADNORM}

\begin{figure}[!htbp]
    \centering
    \begin{subfigure}[b]{0.49\textwidth}
        \centering
        \includegraphics[width=1\linewidth]{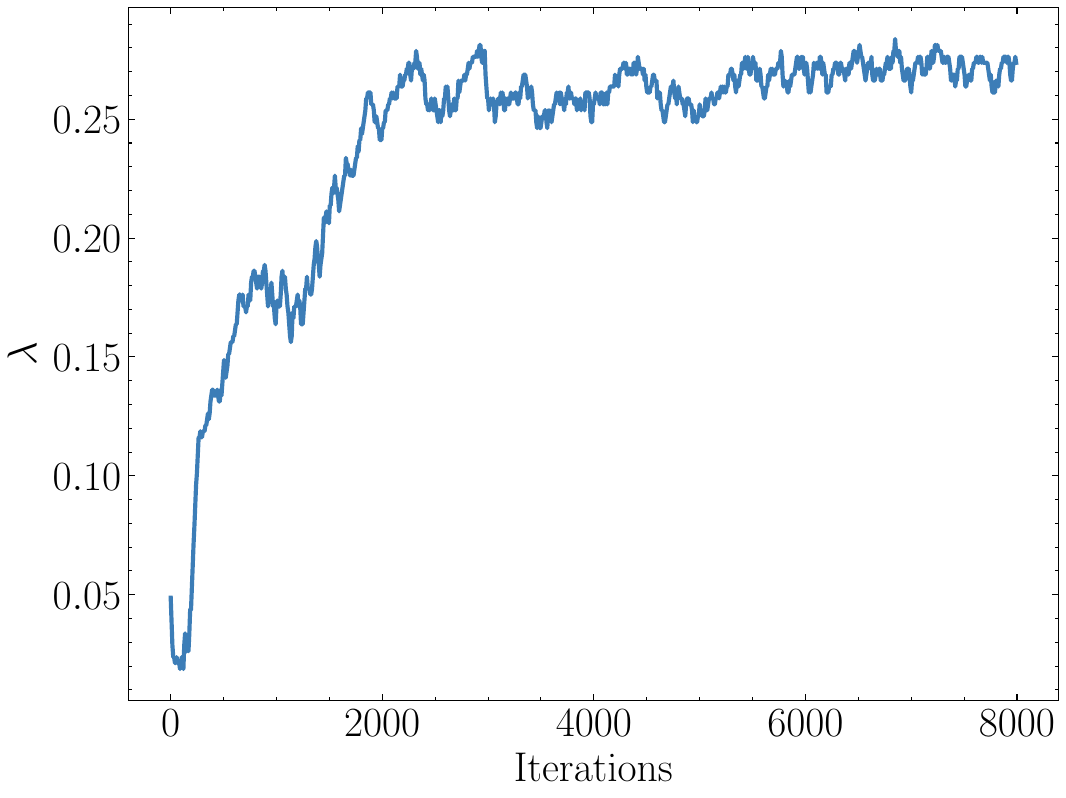} 
        \caption{M3 monthly} 
        \label{fig:Lambda M3}
    \end{subfigure}
    \begin{subfigure}[b]{0.49\textwidth}
        \centering
        \includegraphics[width=1\linewidth]{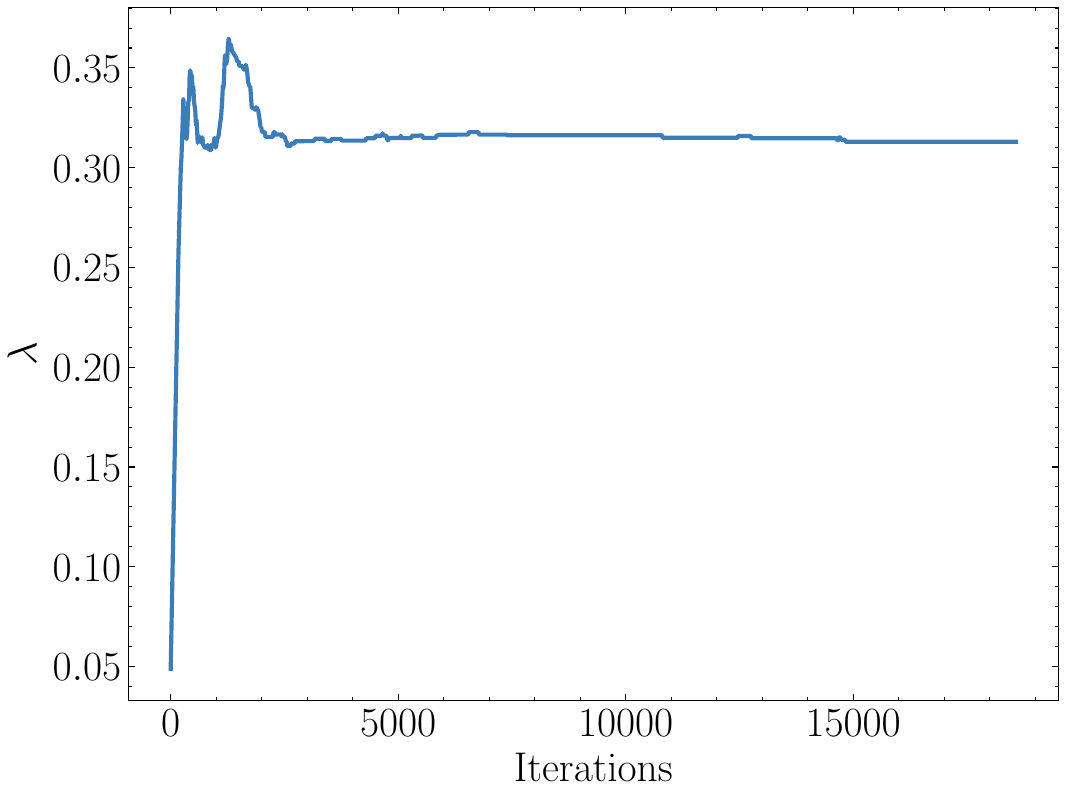} 
        \caption{\textcolor{black}{M4 monthly}}
        \label{fig:Lambda M4}
    \end{subfigure}
    \captionsetup{justification=centering}
    \caption{Evolution of $\lambda_i$ during training using GradNorm.}
    \label{fig:Lambda}
\end{figure}

Recall that the GradNorm algorithm aims to balance the training rates of the different tasks during training. Figure~\ref{fig:Lambda} shows the evolution of $\lambda_i$ during training on the M3 and M4 monthly data sets when using GradNorm. 

For the M3 data set, Figure~\ref{fig:Lambda M3} shows that $\lambda_i$ decreases at the beginning of training and then gradually increases after several iterations.
\textcolor{black}{This pattern 
suggests that forecast stability is initially learned more quickly than forecast accuracy, as $\lambda_i$ decreases to prioritize forecast accuracy early in training to balance the training rates.}
\textcolor{black}{Following this initial learning phase,} 
$\lambda_i$ gradually increases and stabilizes around $0.27$,
\textcolor{black}{which is substantially higher than the static values of $0.025$ and $0.075$ for N-BEATS-S low and high, respectively.}
This supports our hypothesis regarding the usefulness of DLW methods: the model needs to achieve reasonable accuracy before it also considers forecast stability.
However, for the M4 data set, this same behavior is not visible in Figure~\ref{fig:Lambda M4}:
$\lambda_i$ decreases only after the first iteration and then quickly increases to
\textcolor{black}{around 0.31},
where it stabilizes.
An additional experiment showed that lowering the learning rate for the M4 data set results in a similar evolution of $\lambda_i$ as observed for the M3 data set, whereas with the higher learning rate, a reasonable accuracy is already achieved after the first iteration.

\subsubsection{Weighted GCosSim} \label{SEC:DISCUSSION_WGCOSSIM}

\begin{figure}[!htbp]
    \centering
    \begin{subfigure}[b]{0.485\textwidth}
        \centering
        \includegraphics[width=1\linewidth]{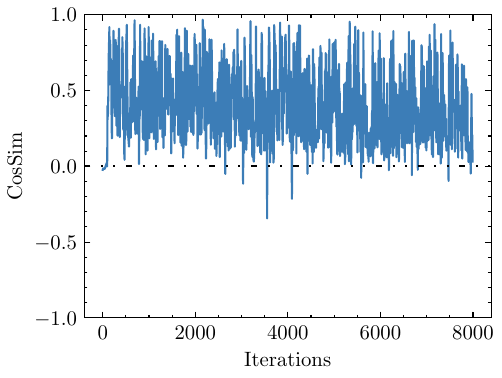} 
        \caption{{M3 monthly}} 
        \label{fig:cosine M3}
    \end{subfigure}
    \begin{subfigure}[b]{0.505\textwidth}
        \centering
        \includegraphics[width=1\linewidth]{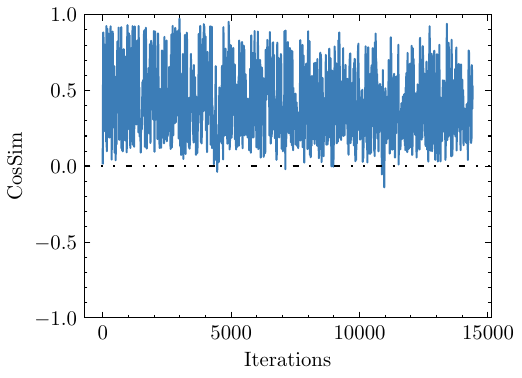} 
        \caption{M4 monthly} 
        \label{fig:cosine M4}
    \end{subfigure}
    \captionsetup{justification=centering}
    \caption{Evolution of cosine similarity between $\vb{g}_\text{error}^i$ and $\vb{g}_\text{instability}^i$ during training using Weighted GCosSim.}
    \label{fig:cosine}
\end{figure}

Figure~\ref{fig:cosine}
shows the evolution of the cosine similarity between $\vb{g}_\text{error}^i$ and $\vb{g}_\text{instability}^i$
during training with Weighted GCosSim on the M3 and M4 monthly data sets.
There are two important observations to discuss.
First, for both data sets, the cosine similarity is generally greater than zero throughout training.
This confirms that optimizing for forecast accuracy and forecast stability can be considered related tasks (after all, if a model generates perfect zero-error forecasts, these forecasts will also be perfectly stable by definition).
Consequently, forecast instability is (almost) consistently taken into account during training with Weighted GCosSim.
The fact that the cosine similarity is substantially less than one for most of the training may explain why regular GCosSim leads to poor accuracy performance.
Since regular GCosSim essentially adds the gradients for both tasks in each iteration without considering the degree of similarity between them, it places too much emphasis on forecast instability given our goal of improving stability without sacrificing accuracy.
Second, for the M3 data set, Figure~\ref{fig:cosine M3} shows that the cosine similarity is close to zero at the start of training, indicating that accuracy and stability are initially unrelated according to cosine similarity.
As training progresses, the cosine similarity becomes positive, suggesting that these tasks become related after this initial phase.
This aligns with the observed evolution of $\lambda_i$ for GradNorm: both algorithms prioritize optimizing forecast accuracy at the start of training before considering forecast instability.
However, as with GradNorm, this behavior is not visible for the M4 data set (see Figure~\ref{fig:cosine M4}).
This can again be attributed to the higher learning rate used for M4,
which leads to a reasonable accuracy 
after just the first iteration.

\textcolor{black}{\subsubsection{AuxiNash}
\label{SEC:DISCUSSION_AUXINASH}}

\begin{figure}[!htbp]
    \centering
    \begin{subfigure}[b]{0.49\textwidth}
        \centering
        \includegraphics[width=1\linewidth]{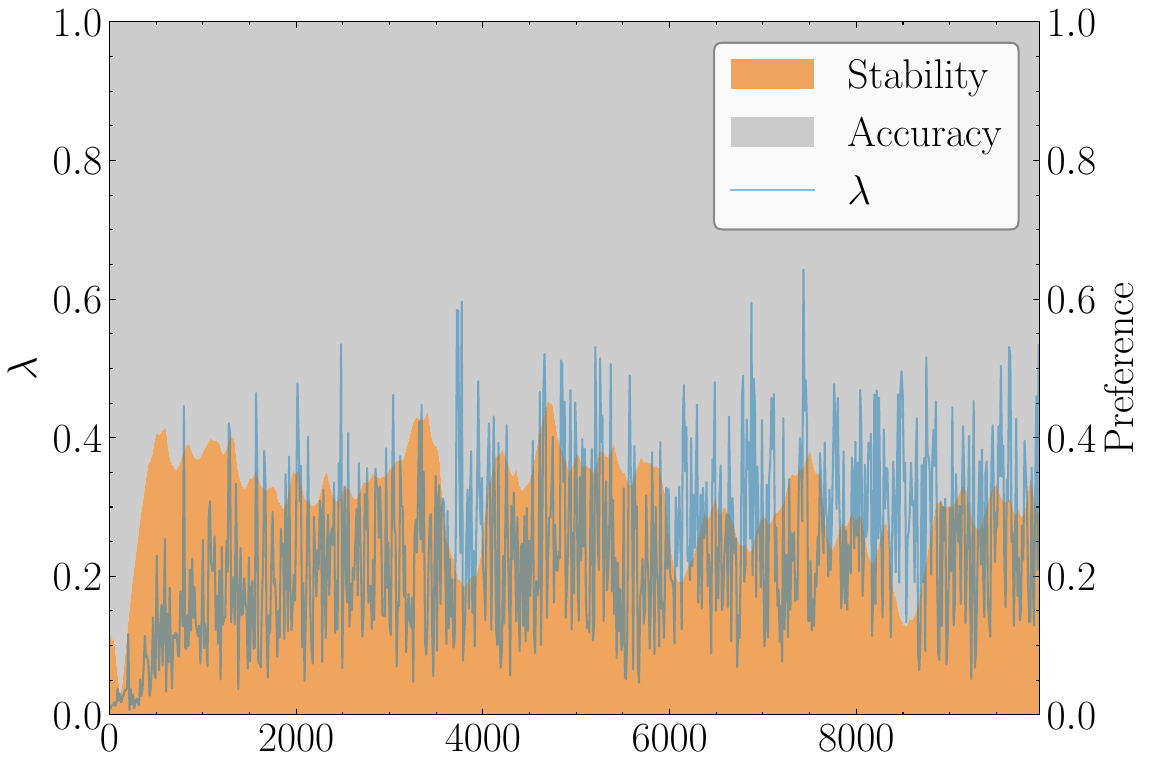} 
        \caption{M3 monthly} 
        \label{fig:preference M3}
    \end{subfigure}
    \begin{subfigure}[b]{0.49\textwidth}
        \centering
        \includegraphics[width=1\linewidth]{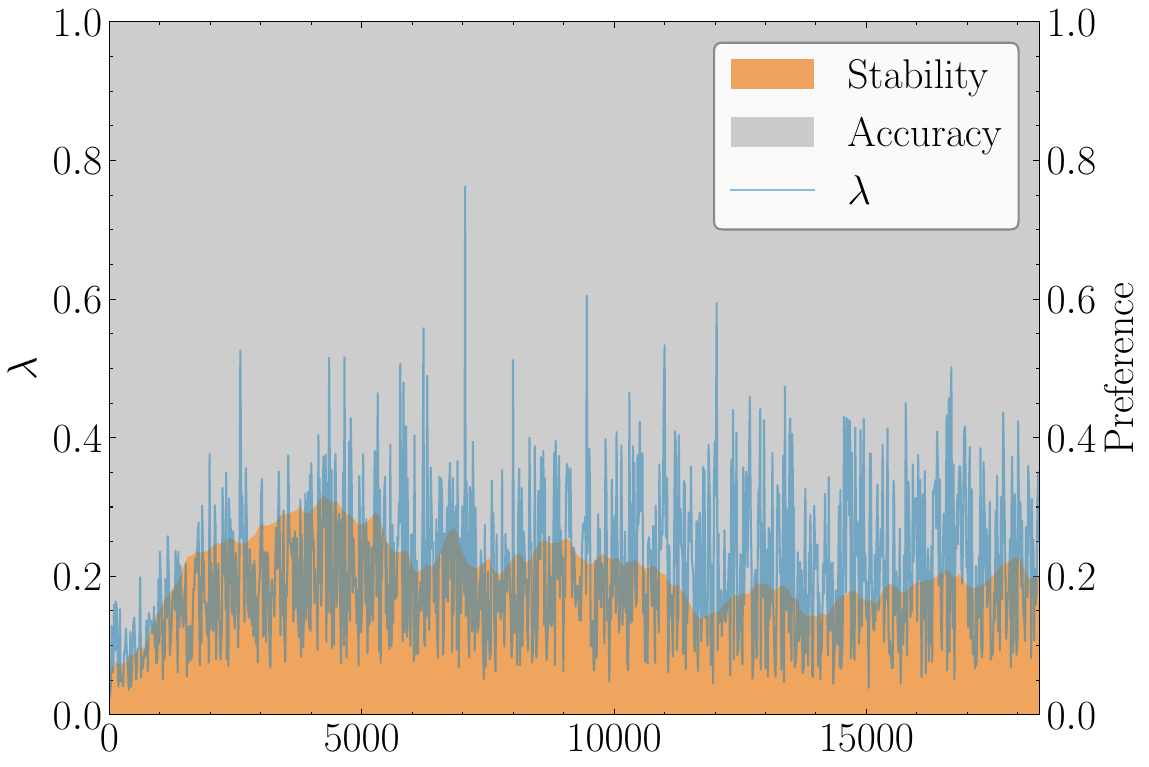} 
        \caption{M4 monthly} 
        \label{fig:preference M4}
    \end{subfigure}
    \captionsetup{justification=centering}
    \caption{\textcolor{black}{Evolution of learned task preferences and $\lambda_i$ during training using AuxiNash.}}
    \label{fig:preference}
\end{figure}

\textcolor{black}{Recall that the AuxiNash algorithm dynamically learns task preferences during training, determining how the bargaining power assigned to stability and accuracy evolves throughout training. These task preferences are subsequently used to calculate $\lambda_i$ (by approximating the solution to the asymmetric bargaining game).
The evolution of the preferences for stability and accuracy, as well as the evolution of $\lambda_i$,
is visualized in Figure~\ref{fig:preference}.}

\textcolor{black}{Similar to GradNorm and Weighted GCosSim, AuxiNash prioritizes accuracy in the first learning iterations for the M3 data set. After this initial learning phase, the preference fluctuates between 0.20 and 0.40 for M3 and between 0.15 and 0.30 for M4, confirming that the algorithm finds that
taking into account stability 
aids in
improving forecast accuracy.}

\subsubsection{TARW} \label{sec:TARW}

\begin{figure}[!htbp]
    \centering
    \begin{subfigure}[b]{0.49\textwidth}
        \centering
        \includegraphics[width=\linewidth]{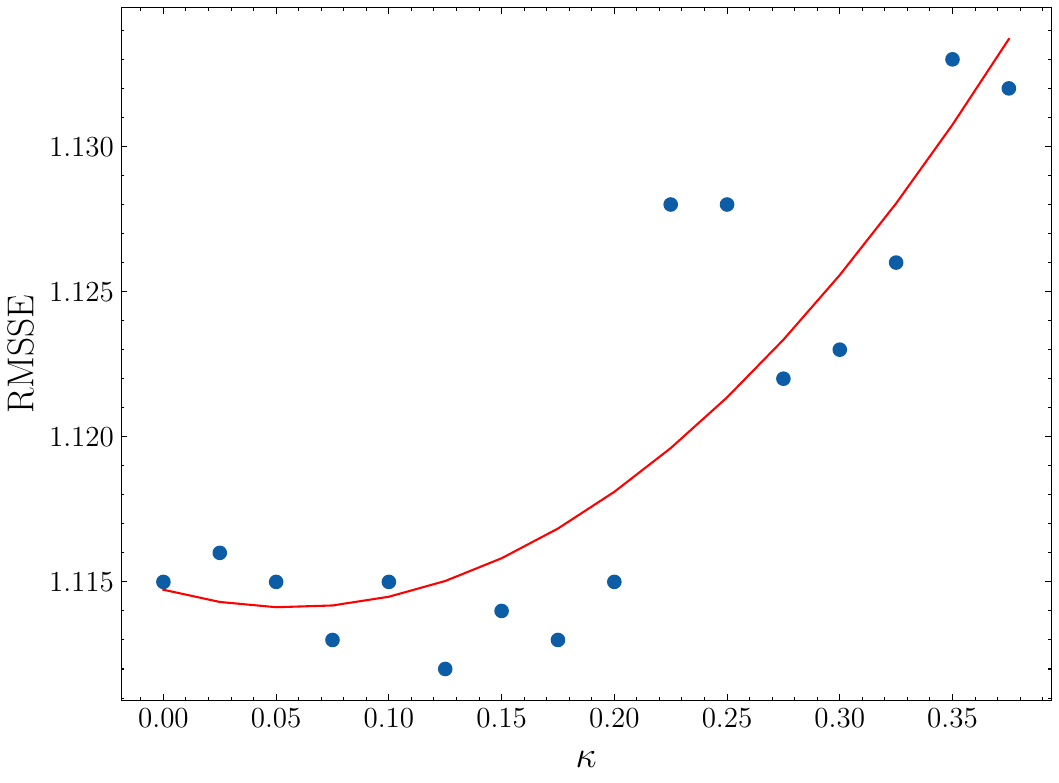} 
        \caption{Accuracy for M3 monthly}
        \label{fig:RMSSE cap M3}
    \end{subfigure}
    \hfill
    \begin{subfigure}[b]{0.475\textwidth}
        \centering
        \includegraphics[width=\linewidth]{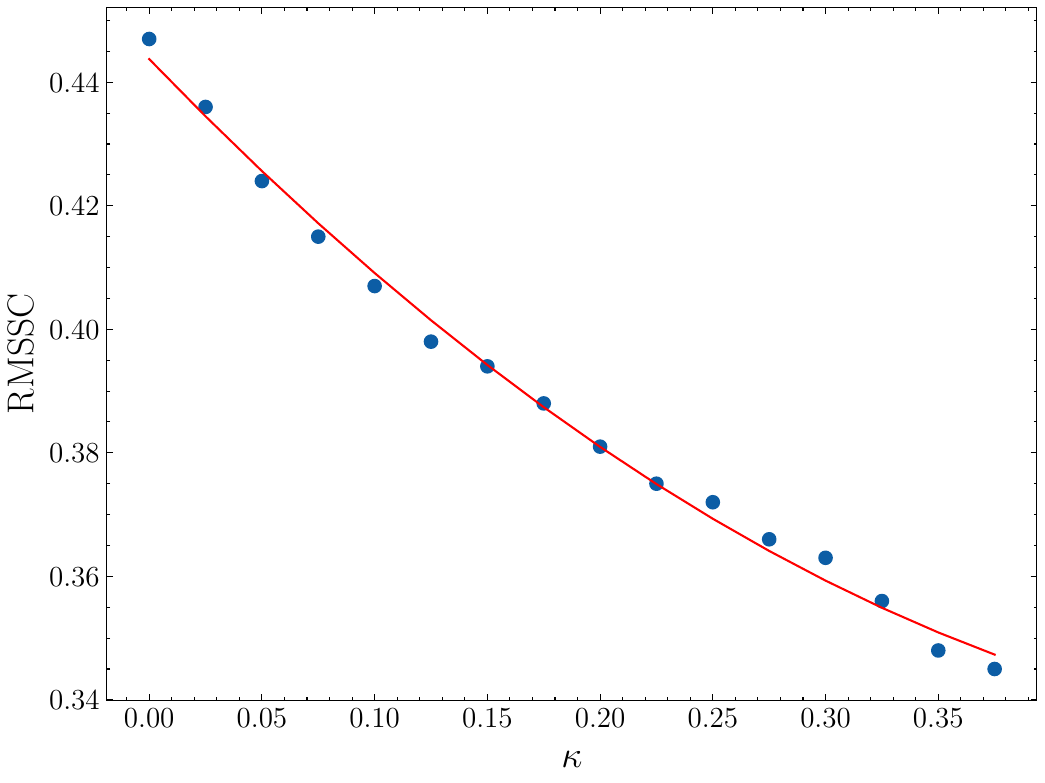} 
        \caption{Stability for M3 monthly} 
        \label{fig:RMSSC cap M3}
    \end{subfigure}
    \hfill
    \begin{subfigure}[b]{0.49\textwidth}
        \centering
        \includegraphics[width=\linewidth]{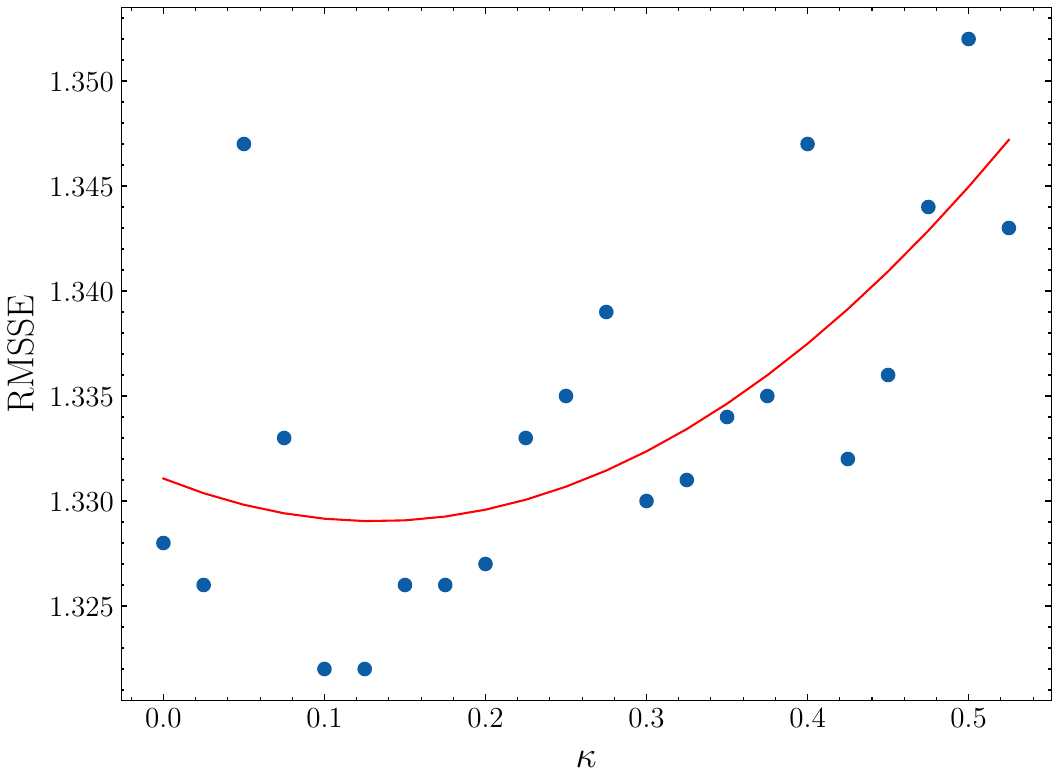} 
        \caption{Accuracy for M4 monthly}
        \label{fig:RMSSE cap M4}
    \end{subfigure}
    \hfill
    \begin{subfigure}[b]{0.485\textwidth}
        \centering
        \includegraphics[width=\linewidth]{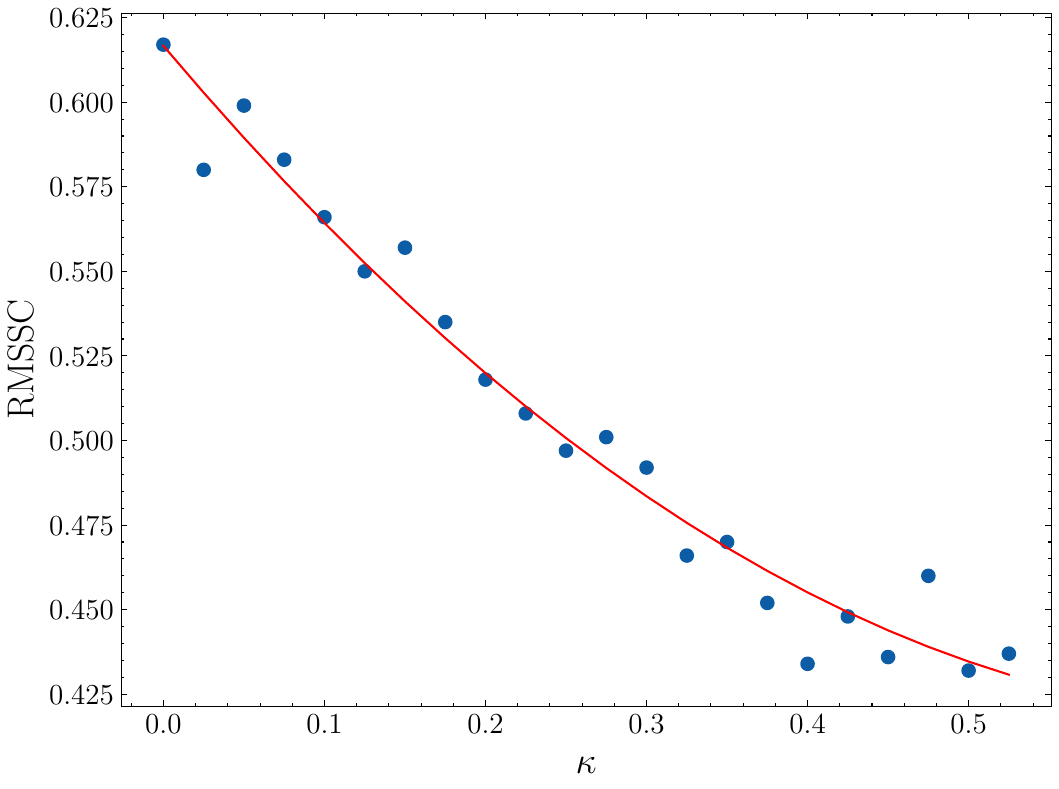} 
        \caption{Stability for M4 monthly} 
        \label{fig:rmssc cap M4}
    \end{subfigure}
    \captionsetup{justification=centering}
    \caption{\textcolor{black}{Validation accuracy (RMSSE) and stability (RMSSC) for TARW as a function of its hyperparameter $\kappa$.}
    }
    \label{fig:cap M3 and M4}
\end{figure}

Figure~\ref{fig:cap M3 and M4} illustrates how 
\textcolor{black}{(smoothed; see Section~\ref{SEC:HYPERPARS})}
validation accuracy (\textcolor{black}{RMSSE}) and stability (\textcolor{black}{RMSSC}) for TARW vary as a function of the hyperparameter $\kappa$ for both the M3 and M4 monthly data sets.
As $\kappa$ increases, a greater average weight is assigned to forecast instability during training.
The impact of $\kappa$ on \textcolor{black}{RMSSE and RMSSC} is similar across both data sets, following trends consistent with those observed for a static $\lambda$ in \citet{vanbelle2023}.
While \textcolor{black}{RMSSC} shows an almost linear decrease with increasing $\kappa$,
\textcolor{black}{RMSSE} initially decreases to a minimum and then increases as $\kappa$ continues to rise (supporting the idea that incorporating forecast instability in training can act as a regularization mechanism).
As discussed in Section~\ref{SEC:HYPERPARS},
the minimum validation \textcolor{black}{RMSSE} is used to select $\kappa$
\textcolor{black}{for TARW low,}
resulting in 
\textcolor{black}{$\kappa = 0.125$}
for both data sets.
\textcolor{black}{This} choice leads to 
\textcolor{black}{$\mathbb{E}[\lambda_i] > \lambda$
for N-BEATS-S low,}
which may (partly) explain the further improvement in forecast stability
\textcolor{black}{without harming forecast accuracy.
The same holds true for TARW high and N-BEATS-S high on the M3 data set, but the relationship does not hold 
on the M4 data set. 
However, it is important to note that N-BEATS-S high for M4 uses a relatively large $\lambda$, resulting in a substantial boost in stability, though at the cost of reduced accuracy in terms of average RMSSE.}
Additionally, we conjecture that TARW, as the stochastic version of static loss weight tuning (see Section~\ref{SEC:METHODOLOGY}),
\textcolor{black}{may outperform}
the latter because it can explore the loss space more effectively and may better escape local optima due to its stochastic nature.

\section{Conclusions and future research} \label{SEC:CONCLUSION}

Rolling origin forecast instability can incur costs, as updates to forecasts based on new data may require changes to plans that rely on these forecasts as inputs.
\citet{vanbelle2023} propose a methodology to optimize global neural point forecasting models from both a forecast accuracy and stability perspective, aiming to improve forecast stability while maintaining accuracy.
They apply this methodology to extend the N-BEATS method \citep{oreshkin2020nbeats}, resulting in a new method called N-BEATS-S.

In this paper, we explore the potential of using DLW methods to train N-BEATS-S,
with the goal of further improving forecast stability while either improving or at least maintaining forecast accuracy compared to N-BEATS-S with static
loss weights.
To this end, we use existing DLW algorithms and propose TARW, a variant of RW \citep{lin2022reasonable}, specifically tailored to align with our objective.
Our empirical results on the M3 and M4 monthly data sets demonstrate that training N-BEATS-S using certain DLW methods can outperform N-BEATS-S with static
loss weights in terms of further improving forecast stability without a significant trade-off in accuracy.
\textcolor{black}{If forecast instability causes (non-technical) forecast consumers to lose trust in the forecasting system, potentially leading to unwarranted judgmental adjustments to the algorithmically obtained forecasts, 
it may be advisable to select a forecasting method that is Pareto efficient but trades some accuracy for a larger improvement in stability (e.g., selecting GradNorm based on the Pareto frontiers shown in Figure~\ref{fig:Pareto plots}).
Using a trusted system that generates 
more stable algorithmic forecasts could yield a better outcome than relying on a system that generates slightly more accurate but less stable forecasts, which are then judgmentally adjusted, potentially reducing forecast accuracy \citep{petropoulos2022forecasting}.}

The motivation for using DLW methods stems from the hypothesis that 
forecast accuracy should be prioritized in the early stages of training,
with forecast instability being addressed only after a reasonable level of accuracy has been achieved.
This approach may lead to better results by enabling a more targeted exploration of the loss space, more closely aligned with our goal.
Our \textcolor{black}{analyses of}
GradNorm (see Section~\ref{SEC:DISCUSSION_GRADNORM}),
Weighted GCosSim (see Section~\ref{SEC:DISCUSSION_WGCOSSIM}),
\textcolor{black}{and AuxiNash (see Section~\ref{SEC:DISCUSSION_AUXINASH})}
support this hypothesis.
Additionally, training N-BEATS-S using TARW also yields favorable results, even outperforming the aforementioned DLW methods in terms of accuracy.
While TARW does not directly prioritize accuracy in the early stages of training but instead randomly samples loss weights from a uniform distribution $U(0,\kappa)$, where $\kappa \in (0, 1]$ is a tunable hyperparameter, we believe this stochastic variant of static loss weight tuning is effective because it allows for a more comprehensive exploration of the loss space and may better escape local optima due to its stochastic nature.

\textcolor{black}{While the computational complexity during inference remains the same as static loss weighting for all DLW methods, a drawback is the increased computational complexity during training.
Specifically, most approaches
involve further processing of the individual task gradients (e.g., in NashMTL and AuxiNash, the individual task gradients are used to approximate the (generalized) Nash bargaining solution).
RW and TARW, on the other hand, only require sampling from a uniform distribution per training iteration, making them only marginally more computationally expensive than static loss weighting.
By also taking into account the results for TARW, its ease of implementation, and the fact that it requires tuning just one hyperparameter---similar to static loss weight tuning---we recommend using at least TARW instead of static loss weights.}

One limitation of this work is that we only incorporate forecast stability for adjacent forecasting origins during optimization \citep[as in][]{vanbelle2023}.
While it is intuitive that this approach would also improve stability for non-adjacent forecasting origins, the effectiveness of directly accounting for instability with respect to non-adjacent origins during optimization should be explored in future work.
Additionally, to further validate our findings, the benefits of training N-BEATS-S using DLW methods should be tested on a broader range of data sets.
\textcolor{black}{Furthermore, the effectiveness of DLW methods in optimizing the composite loss function
proposed by \citet{vanbelle2023}---or a similar composite loss function that combines forecast error and forecast instability---should be evaluated for neural network architectures other than N-BEATS.}
Another limitation of our study is the ensemble size.
As explained in Section~\ref{SEC:FORECASTING_METHODS}, we use ensembles consisting of only five models, whereas \citet{oreshkin2020nbeats} used a total of 180 models for each of their final N-BEATS ensembles.
Investigating how ensemble size and type affect forecast accuracy and stability is an interesting direction for future research.

\textcolor{black}{Additionally}, another promising area for future research is the application of TARW in other MTL settings.
Given that \citet{lin2022reasonable} demonstrated the superiority of RW over equal weighting, 
it is plausible that TARW could also outperform static loss weight tuning in other MTL problems due to its stochastic nature.
\textcolor{black}{In this regard, similar to the conclusion of \cite{lin2022reasonable} that RW should be considered a strong baseline in MTL problems, 
and based on the performance reported in this work,
we advocate for the use of TARW as a baseline DLW method in auxiliary learning settings.}
Finally, a broader direction for future research concerns incorporating forecast stability into different modeling approaches or pipelines,
such as the post-processing technique for stabilizing point forecasts proposed by \citet{godahewa2023forecast}
or the N-BEATS variant to stabilize Gaussian probabilistic forecasts by \citet{vanbelle2024}.
Of particular interest is the incorporation of forecast stability into tree-based methods like LightGBM \citep{ke2017lightgbm} due to their widespread adoption in the time series forecasting field, as evidenced in the M5 competition \citep{makridakis2022m5}.

\section*{Acknowledgements}

Withheld for peer review.


\section*{Declaration of generative AI and AI-assisted technologies in the writing process}

During the preparation of this work the authors used ChatGPT in order to improve language and readability.
After using this tool/service, the authors reviewed and edited the content as needed and take full responsibility for the content of the publication.


\bibliographystyle{newapa}

\appendix

\setcounter{section}{0}
\renewcommand{\thesection}{Appendix \Alph{section}}
\renewcommand{\thefigure}{\Alph{section}.\arabic{figure}}

\section{\textcolor{black}{Results in terms of sMAPE and sMAPC}}
\label{sec:results_appendix}
\setcounter{figure}{0}
\setcounter{table}{0}

\textcolor{black}{Table~\ref{tab:results_appendix} summarizes the results
in terms of sMAPE to evaluate forecast accuracy:
\begin{equation} \label{EQ:SMAPE}
    \text{sMAPE}(\mathbf{\hat{y}}_{h|t}^j,\mathbf{y}_{h|t}^j) = \frac{200}{h} \sum_{i=1}^{h} \frac{|y_{t+i}-\hat{y}_{t+i|t}|}{|y_{t+i}|+|\hat{y}_{t+i|t}|},
\end{equation}
and sMAPC to evaluate forecast stability:
\begin{equation} \label{EQ:SMAPC}
    \text{sMAPC}(\mathbf{\hat{y}}_{h|t}^j,\mathbf{\hat{y}}_{h|t-1}^j) = 
    \frac{200}{(h-1)} \sum_{i=1}^{h-1} \frac{|\hat{y}_{t+i|t-1}-\hat{y}_{t+i|t}|}{|\hat{y}_{t+i|t-1}|+|\hat{y}_{t+i|t}|}.
\end{equation}}

\begin{figure}[!htbp]
\centering
\begin{subfigure}[b]{0.49\textwidth}
\centering
\includegraphics[width=1\linewidth]{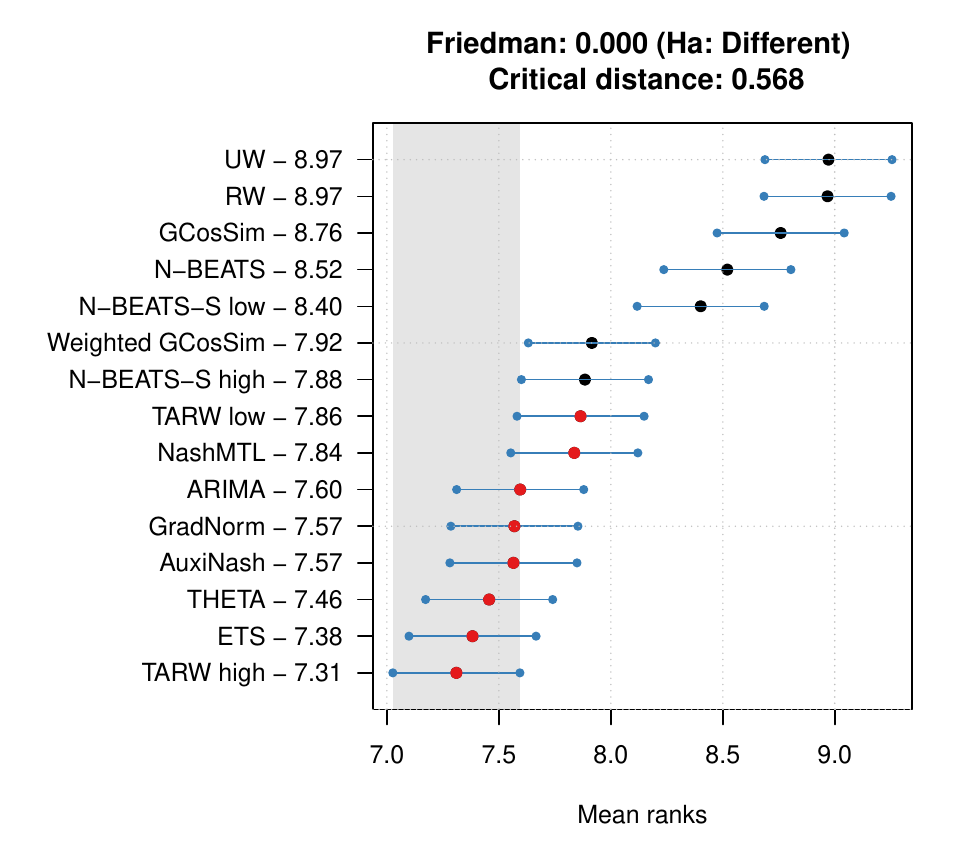}
\caption{sMAPE ranking} 
\label{fig:MCB M3 ACC smape}
\end{subfigure}
\begin{subfigure}[b]{0.49\textwidth}
\centering
\includegraphics[width=1\linewidth]{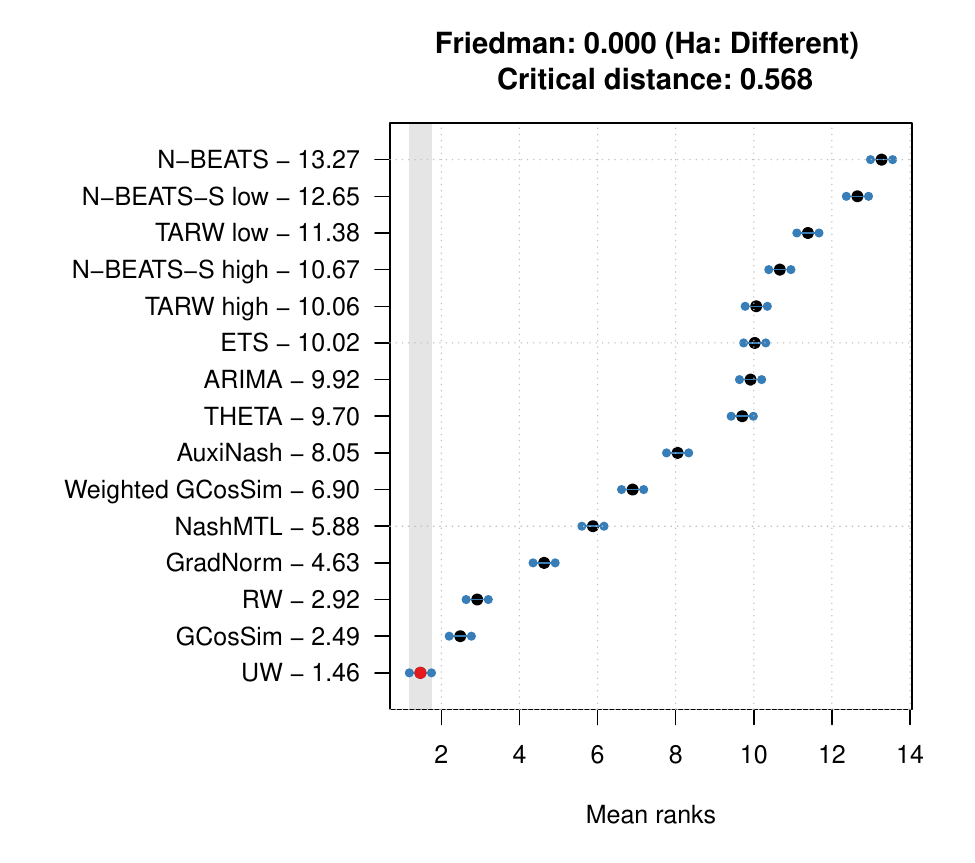} 
\caption{sMAPC ranking}
\label{fig:MCB M3 stab smapc}
\end{subfigure}
\captionsetup{justification=centering}
\caption{\textcolor{black}{MCB results for the M3 monthly data set. 
Lower is better. 
If two intervals overlap, there is no statistically significant difference between the corresponding methods. }
\label{fig:MCB M3 app}}
\end{figure}

\begin{figure}[!htbp]
\centering
\begin{subfigure}[b]{0.49\textwidth}
\centering
\includegraphics[width=1\linewidth]{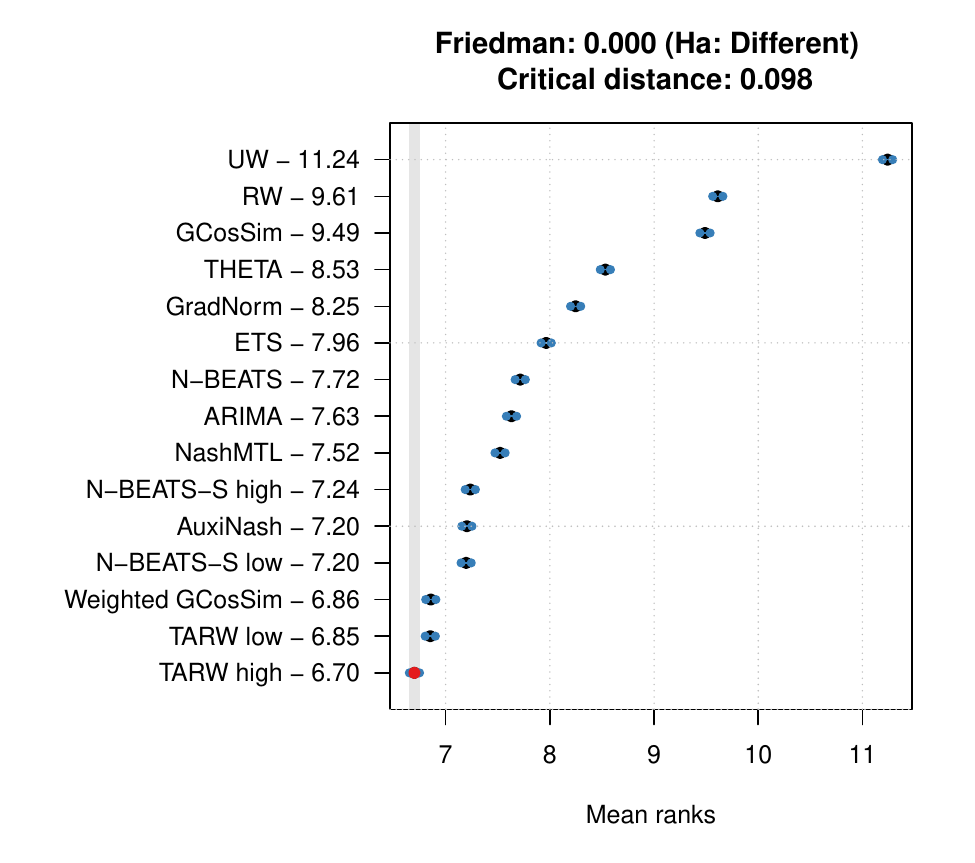}
\caption{sMAPE ranking} 
\label{fig:MCB M4 ACC smape}
\end{subfigure}
\begin{subfigure}[b]{0.49\textwidth}
\centering
\includegraphics[width=1\linewidth]{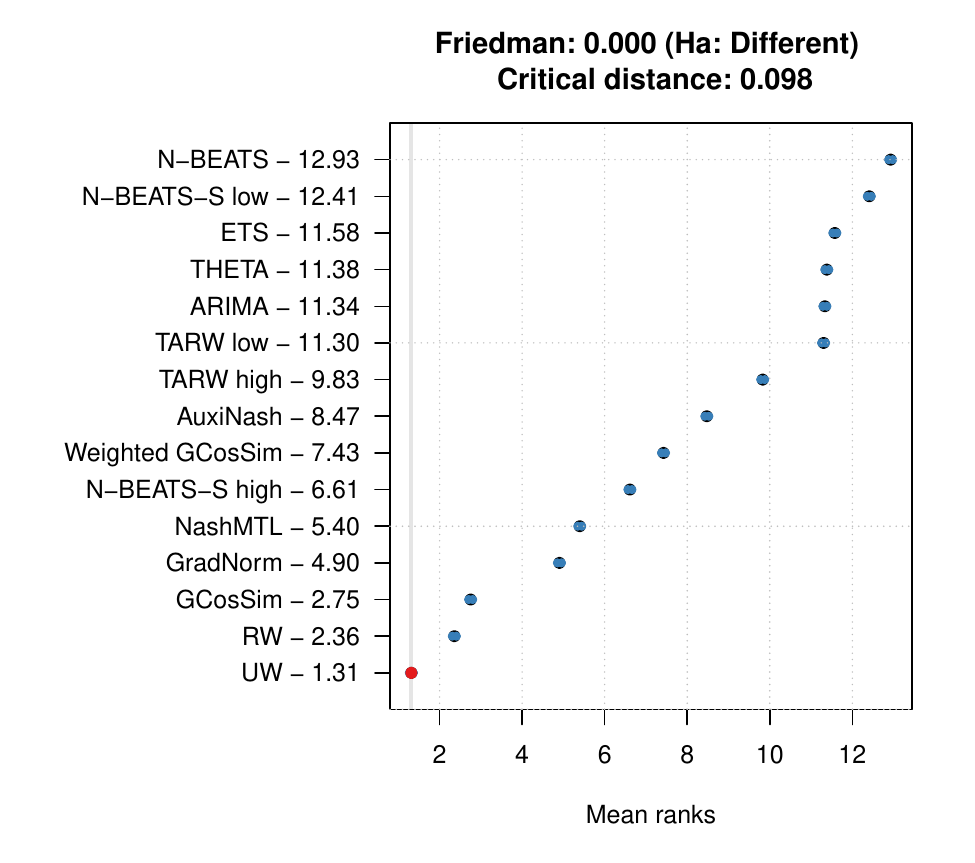} 
\caption{sMAPC ranking}
\label{fig:MCB M4 stab smapc}
\end{subfigure}
\captionsetup{justification=centering}
\caption{\textcolor{black}{MCB results for the M4 monthly data set. 
Lower is better.
If two intervals overlap, there is no statistically significant difference between the corresponding methods.}
\label{fig:MCB M4 app}}
\end{figure}

\textcolor{black}{The results of MCB tests based on sMAPE and sMAPC rankings are shown in Figures~\ref{fig:MCB M3 app} and \ref{fig:MCB M4 app} for the M3 and M4 data sets, respectively.}

\begin{table}[!htbp]
\centering
\begin{tabular}{lccccc}
\hline
& \multicolumn{2}{c}{{M3 monthly}} & & \multicolumn{2}{c}{{M4 monthly}} \\
\cline{2-3} \cline{5-6}
& {sMAPE} & {sMAPC} & & {sMAPE} & {sMAPC} \\
\hline
\multicolumn{1}{l}{N-BEATS} & 11.44 & 3.65 & & \textbf{9.12} & 3.88 \\
\hline
\multicolumn{1}{l}{\textcolor{black}{N-BEATS-S low}} & 11.43 & 3.45 & & \textbf{9.12} & 3.69 \\
\multicolumn{1}{l}{\textcolor{black}{N-BEATS-S high}} & 11.40 & 3.07 & & {9.24} & 2.22 \\
\hdashline
\multicolumn{1}{l}{GradNorm} & 11.47 & 1.63 & & \textcolor{black}{9.37} & \textcolor{black}{1.89} \\
\multicolumn{1}{l}{UW} &11.62 &\textbf{0.92} & &10.56 &\textbf{0.84} \\
\multicolumn{1}{l}{\textcolor{black}{NashMTL}} & 11.42 & 1.82 & & 9.27 & 2.06 \\
\multicolumn{1}{l}{RW} & 11.64 & \underline{1.06} & & 9.76 & \underline{1.22} \\
\hdashline
\multicolumn{1}{l}{GCosSim} & 11.59 & {1.05} & & 9.70 & 1.28 \\
\multicolumn{1}{l}{Weighted GCosSim} & 11.41 & 2.07 & & 9.18 & 2.40 \\

\multicolumn{1}{l}{\textcolor{black}{AuxiNash}} & 11.39 & 2.28 & & 9.15 & 2.64 \\
\multicolumn{1}{l}{\textit{\textcolor{black}{TARW low}}} & 11.42 & 3.22 & & $\underline{9.13}$ & 3.36 \\
\multicolumn{1}{l}{\textit{\textcolor{black}{TARW high}}} & 11.40 & 2.97 & & $\underline{9.13}$ & 3.04 \\
\hline
\multicolumn{1}{l}{ETS} & \underline{11.34} & 3.21 & & 9.98 & 4.38 \\
\multicolumn{1}{l}{ARIMA} & 11.70 & 3.16 & & 9.78 & 4.15 \\
\multicolumn{1}{l}{THETA} & \textbf{11.28} & 2.96 & & 10.07 & 3.80 \\
\hline
\end{tabular}
\captionsetup{justification=centering}
\caption{\textcolor{black}{Forecast accuracy and stability performance on the
test sets. 
Lower is better.\\
The minimum value per column is highlighted in bold.}}
\label{tab:results_appendix}
\end{table}

\end{document}